\documentclass[11pt]{article}

\usepackage[final]{acl}
\usepackage{times}
\usepackage{latexsym}

\usepackage[T1]{fontenc}

\usepackage[utf8]{inputenc}

\usepackage{microtype}

\usepackage{inconsolata}

\usepackage{graphicx}

\usepackage{tcolorbox}
\usepackage{booktabs}
\usepackage{multirow}
\usepackage{tabularx}

\usepackage{times}
\usepackage{helvet}
\usepackage{courier}
\usepackage{xcolor}
\usepackage{colortbl}
\usepackage{amssymb}

\usepackage[table]{xcolor}
\usepackage{pifont}
\newcommand{\cmark}{\ding{51}} 
\newcommand{\xmark}{\ding{55}} 

\tcbuselibrary{breakable}

\title{\textsc{ElaipBench}: A Benchmark for Expert-Level Artificial \\ Intelligence Paper Understanding}

\author{
Xinbang Dai$^{\dagger*}$ 
\quad Huikang Hu$^{\dagger*}$ 
\quad Yongrui Chen$^\dagger$ 
\quad Jiaqi Li$^\dagger$ 
\quad Rihui Jin$^\dagger$\\
\textbf{Yuyang Zhang$^\diamond$
\quad Xiaoguang Li$^\diamond$ 
\quad Lifeng Shang$^\diamond$ 
\quad Guilin Qi$^\dagger$} \\
$^\dagger$Southeast University  \quad  $^\diamond$Noah's Ark Lab \\
\texttt{\{xbdai, huikanghu\}@seu.edu.cn} \\
\phantom{\thanks{Equal contribution.}}
}

\begin{document}
\maketitle

\begin{abstract}
While large language models (LLMs) excel at many domain-specific tasks, their ability to deeply comprehend and reason about full-length academic papers remains underexplored. Existing benchmarks often fall short of capturing such depth, either due to surface-level question design or unreliable evaluation metrics. To address this gap, we introduce \textsc{ElaipBench}, a benchmark curated by domain experts to evaluate LLMs’ comprehension of artificial intelligence (AI) research papers. Developed through a game-theoretic, adversarial annotation process, \textsc{ElaipBench} features 403 multiple-choice questions from 137 papers. It spans three difficulty levels and emphasizes non-trivial reasoning rather than shallow retrieval. Our experiments show that the best-performing LLM achieves an accuracy of only 39.95\%, far below human performance. Moreover, we observe that frontier LLMs equipped with a thinking mode or a retrieval-augmented generation (RAG) system fail to improve final results—even harming accuracy due to overthinking or noisy retrieval. These findings underscore the significant gap between current LLM capabilities and genuine comprehension of academic papers.
\footnote{The benchmark is publicly available at \url{https://huggingface.co/datasets/KangKang625/ELAIPBench}. Building on this work, we organized the \emph{CCKS-2025 Complex Question Answering Competition for AI Papers} using a subset of \textsc{ElaipBench}. The final rankings are accessible at \url{https://tianchi.aliyun.com/competition/entrance/532359/rankingList}.}
\end{abstract}

\section{Introduction}

In recent years, large language models (LLMs) have made remarkable advancements in acquiring domain-specific knowledge and enhancing reasoning capabilities. These models encapsulate extensive expertise across various fields, enabling performance at or beyond human-expert levels on multiple scientific benchmarks~\cite{achiam2023gpt, lala2023paperqa, laurent2024lab, du2025supergpqa}. Furthermore, LLMs have demonstrated strong reasoning abilities on tasks that require logical deduction~\cite{liu2024deepseek, yang2025qwen3}, including mathematics~\cite{hendrycks2021measuring} and programming~\cite{hui2024qwen2, jain2025livecodebench}. Given that artificial intelligence (AI) research often requires a strong command of both mathematics and algorithmic programming, these advancements have led AI researchers to increasingly employ LLMs as tools for reading, interpreting, and even reviewing academic papers. However, this raises a critical question: \textbf{To what extent can LLMs truly comprehend academic content—specifically, are they capable of deep understanding, learning, and reasoning based on lengthy scholarly texts?}

\begin{figure}[t]
  \centering
  \includegraphics[width=\columnwidth]{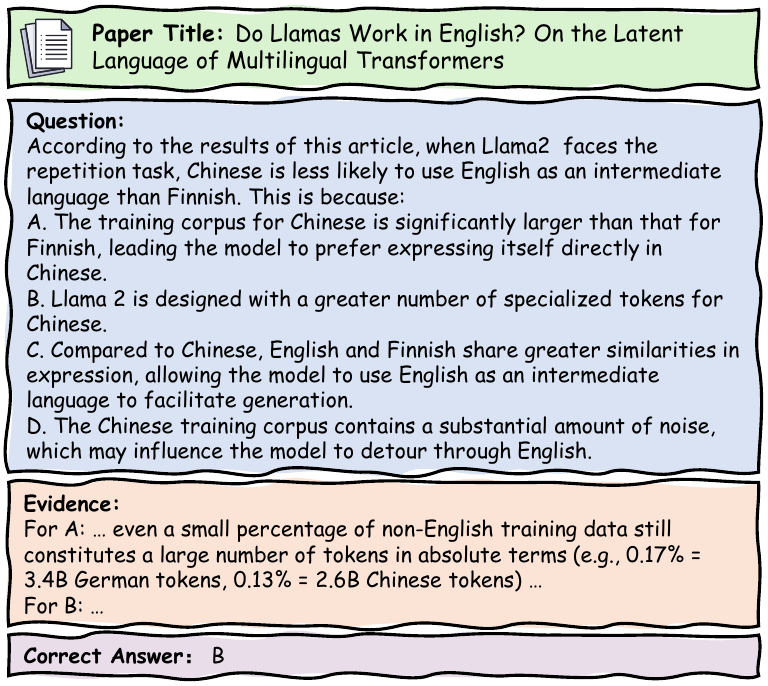}
  \caption{An example from \textsc{ElaipBench}. Option A, while potentially factually correct, is not substantiated by the text. Option C is a deliberately constructed distractor with no direct supporting evidence in the text. Option D is also not explicitly stated in the text, although its latter part is derived from the paper.}
  \label{fig1}
\end{figure}

Unfortunately, existing domain-specific benchmarks fail to adequately assess the deep comprehension of LLMs for academic papers. Some benchmarks focus on content extraction tasks where answers are explicitly stated in the papers~\cite{lala2023paperqa, auer2023sciqa, laurent2024lab}. These answers can be easily retrieved by LLMs due to their strong recall abilities. Others employ open-ended questions requiring the summarization of phenomena or methods described in the papers~\cite{saikh2022scienceqa, lee2023qasa, wan2024sciqag, gui2025acadreason}; however, their evaluation metrics (e.g., ROUGE) fail to accurately and reliably measure fine-grained details or key technical knowledge. Crucially, existing approaches relying on human annotation struggle with quality control—although some incorporate question validation mechanisms~\cite{welbl2017crowdsourcing, bai2024longbench}, crowd-sourced annotation schemes may still produce superficial questions~\cite{kittur2008crowdsourcing}.

To address these limitations, we aim to construct a benchmark with the following features: (1) \textbf{Reliability}: all evaluations should be conducted in the form of multiple-choice questions, ensuring a precise and standardized format; (2) \textbf{Difficulty}: the questions should be challenging enough that even experienced researchers cannot answer all of them correctly in a short time; (3) \textbf{High Quality}: all questions must be answerable, and each must be accompanied by evidence-supported answers to ensure both validity and interpretability.

With the above goal in mind, we introduce \textsc{ElaipBench}, a benchmark designed to evaluate \textbf{E}xpert-\textbf{L}evel \textbf{A}rtificial \textbf{I}ntelligence \textbf{P}aper understanding. \textsc{ElaipBench} encompasses papers from the domains of machine learning (ML), computer vision (CV), and natural language processing (NLP). It comprises 403 manually crafted multiple-choice questions in both single-answer and multiple-answer formats, thereby ensuring the \textbf{reliability} of evaluation. The benchmark, presented in English, provides each question with its full-length source paper, the correct answers, and the corresponding evidence excerpt from the paper.

To ensure the \textbf{difficulty} and \textbf{high quality} of \textsc{ElaipBench}, we employ a game-theoretic competitive annotation protocol designed to yield challenging questions. We recruit 20 human annotators, each holding at least a master’s degree in computer science and having prior experience in publishing academic papers. Each annotator is assigned one of three roles: \emph{Question Writer}, \emph{Evidence Verifier}, or \emph{Answer Verifier}.The \emph{Question Writer} is responsible for creating data instances. Each instance comprises an English academic paper, a set of questions with corresponding answer choices, the correct answer, and supporting evidence excerpts from the paper. Subsequently, as a preliminary difficulty filter, we use three LLMs for automated review. If any model answers the question correctly, it is discarded as too easy. The filtered questions are assigned to Evidence Verifiers, who answer them using the paper and evidence excerpts. A question is retained only if the \emph{Evidence Verifier} derives the correct answer using the provided evidence. Finally, the \emph{Answer Verifier} attempts to answer the questions based solely on the paper, question, and options. Throughout the process, annotators receive a base salary and compete for performance bonuses based on response time and accuracy.  By adopting this game-theoretic incentive design, we establish \emph{careful annotation} as the Nash equilibrium for all annotators, thereby yielding challenging yet answerable questions. Figure~\ref{fig1} presents an example question from \textsc{ElaipBench}. Its options not only incorporate explicit content from the paper but also require LLMs to perform reasoning based on this content combined with scientific common knowledge. 
Overall, our contributions are summarized as follows:

\begin{itemize}
    \item We introduce \textsc{ElaipBench}, a benchmark containing 403 expert-created questions with three difficulty levels. The questions are precise and unambiguous, and they require deep understanding beyond simple paper retrieval.

    \item We propose a game-theoretic annotation mechanism that uses performance-linked bonuses to generate high-quality questions. Comprehensive evaluations show the benchmark's difficulty: human experts achieve only 48.14\% accuracy within 20 minutes, while the best-performing LLMs reach just 39.95\%. 
    
   \item In our experiments, we observe that LLMs augmented with either reasoning or retrieval capabilities generally underperform non-augmented baselines on \textsc{ElaipBench}. This suggests that current naive augmentation methods may lead to misinterpretations of papers, which highlights an unexplored gap for future research.

\end{itemize}

\section{Related Works}
\subsection{Scientific Comprehension Benchmarks}

To advance the evaluation of LLMs in the domain of scientific comprehension, a variety of benchmark datasets have emerged, with increasing emphasis on task complexity and domain specificity. These benchmarks can be broadly categorized into two groups. The first group either does not require contextual information or relies on only brief text passages. Sorted by publication date, these benchmarks include emrQA~\cite{pampari2018emrqa}, PubMedQA~\cite{jin2019pubmedqa}, ScienceQA~\cite{saikh2022scienceqa}, SciQA~\cite{auer2023sciqa}, SciBench~\cite{wang2023scibench}, ChemLit-QA~\cite{wellawatte2025chemlit}, and HLE~\cite{phan2025humanity}. They primarily assess models' intrinsic knowledge within a specific domain while neglecting the critical ability to comprehend contextual information—an essential skill in academic research or paper review. The second group incorporates full academic papers as context~\cite{lala2023paperqa, skarlinski2024language, bai2024longbench}; however, the associated questions tend to be superficial, enabling LLMs to locate answers through direct text matching rather than requiring synthesis, inference, or critical engagement with the academic paper context.  For more detailed comparisons of existing benchmarks, please refer to Appendix~\ref{dataset_comparsion}.

\subsection{Annotation Methods}


Current academic question answering benchmarks are primarily generated using two annotation approaches. The first relies on crowdsourced human annotation, where data labeling is conducted through pipeline-based workflows and quality control is typically managed by a review group~\cite{lala2023paperqa, bai2024longbench, skarlinski2024language, asai2024openscholar, gui2025acadreason}. This approach, however, heavily depends on the efficiency and expertise of the reviewing team, with limited mechanisms for providing feedback on the review outcomes. The second approach leverages LLMs to automatically generate datasets by equipping them with various specialized tools and prompting frameworks~\cite{lee2023qasa, wan2024sciqag, kim2025autopaperbench, yu2025scicueval}. While scalable, this method is constrained by the inherent comprehension limits of the models, frequently resulting in insufficient diversity and a lack of depth required for resolving complex academic questions. To the best of our knowledge, we are the first to integrate game-theoretic principles into the annotation process for academic QA benchmarks.
\section{Construction of \textsc{ElaipBench}}

In this section, we present the construction process of \textsc{ElaipBench}. First, we define the task. Next, we introduce a game-theoretic annotation mechanism designed to ensure challenging data. Finally, we provide statistical information.

\begin{figure*}[t]
  \centering
  \includegraphics[width=0.88\textwidth]{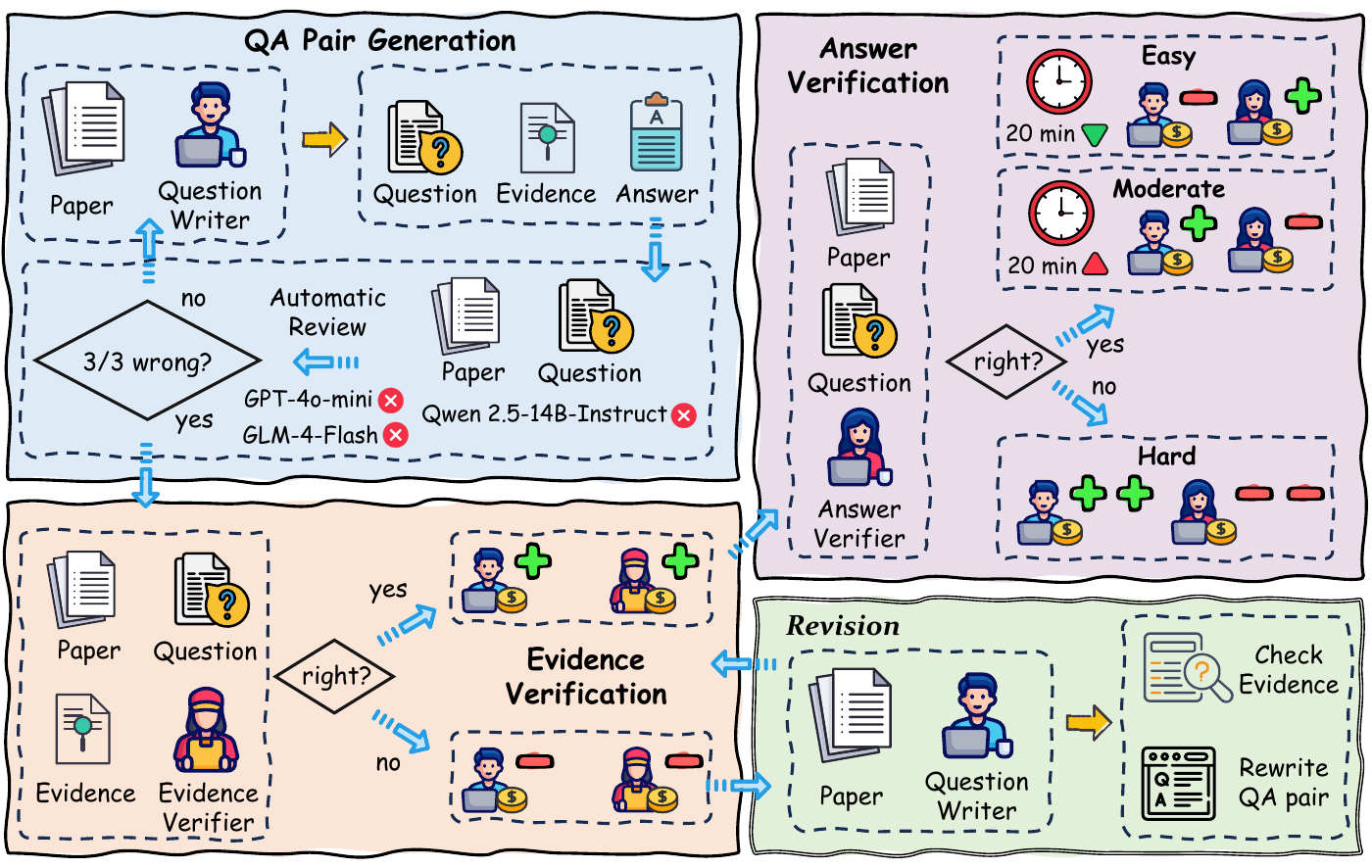}
  \caption{Construction of \textsc{ElaipBench}.}
  \label{fig2}
\end{figure*}

\subsection{Task Overview}
\paragraph{Task Definition.} The task is formally defined as follows: given an academic paper text $p$ from the AI domain, published in a conference or journal, and a multiple-choice question $q$, the model is required to identify all correct answer choices based on the content of the paper. Any response containing incorrect options or failing to include all correct choices receives zero points.

\paragraph{Annotation Team.} To collect high-quality and challenging question-answering (QA) data derived from academic papers, we recruit a team of 20 annotators from top universities. We also develop an online annotation platform designed to facilitate the annotation process. All annotators have previously published papers in the field of AI and are fluent in English. Specifically, the team comprises one professor, one postdoctoral researcher, ten Ph.D. students, and eight master’s students, all specializing in AI. Further details about the annotation platform and team are provided in Appendix~\ref{annotation_details}.

\subsection{Game-Theoretic Annotation Mechanism}
As illustrated in Figure~\ref{fig2}, our annotation mechanism consists of three key stages: QA Pair Generation, Evidence Verification, and Answer Verification. These stages involve three distinct roles (\emph{Question Writer}, \emph{Evidence Verifier}, and \emph{Answer Verifier}), staffed with 10, 4, and 6 annotators, respectively. The roles operate independently yet pursue competing objectives within the workflow. Compensation is structured as a combination of base pay and performance-based bonuses. Annotators receive a base payment of 30 CNY per question they generate. During the verification stages, they can earn additional performance bonuses of 30 CNY (Level-1) or 60 CNY (Level-2), awarded based on the quality of their annotations. The total cost for all data collection and annotation was over 50,000 CNY (\textasciitilde7,000 USD).

\subsubsection{QA Pair Generation}

In the QA Pair Generation stage, each \emph{Question Writer} is required to first upload a long paper from a top conference or journal with which they are thoroughly familiar. The annotation platform employs a PDF-to-text tool such as PyMuPDF\footnote{https://github.com/pymupdf/PyMuPDF} to transform the uploaded file into plain text. This preprocessed text is then automatically checked for length, and any document containing fewer than 4,096 tokens is rejected due to insufficient content for generating challenging questions. Following successful preprocessing, the \emph{Question Writer} creates both single-answer multiple-choice questions (SA-MCQ) and multiple-answer multiple-choice questions (MA-MCQ) according to detailed annotation guidelines. Each SA-MCQ must have exactly one correct answer, while each MA-MCQ must have one or more correct answers. All correct answers must be grounded in factual content from the paper and require corresponding supporting evidence. This supporting evidence must consist of verbatim excerpts from one or more paragraphs of the source paper. Its sufficiency is determined by whether an AI expert could correctly answer the question based solely on this evidence.

Our annotation guidelines are as follows: 
(1) Avoid selecting highly-cited or canonical academic papers to ensure diversity and mitigate the risk of models relying on memorized knowledge.
(2) Questions must be objectively formulated in English, free of subjectivity or ambiguity.
(3) The correct answer must not be obtainable through simple keyword matching or direct retrieval from the text; instead, answering the question should require reasoning, summarization, or integration of multiple pieces of information from different sections of the paper.
(4) Questions and options should not introduce domain-specific terms or abbreviations unless they are explicitly defined or already appear in the paper, as such content may confuse LLMs or introduce unintended ambiguity.
(5) Distractors should be carefully designed to include misleading phrasing or partial truths—statements that align with some aspects of the paper but are ultimately incorrect—thereby increasing the cognitive demand and necessitating careful comparison and deep reasoning.
(6) For SA-MCQ, exactly one option must be correct.
(7) For MA-MCQ, the number of correct options must be either two or three, a constraint designed to ensure a balanced level of difficulty and discourage random guessing. The detailed annotation guidelines are provided in Appendix~\ref{guidelines}.

The \emph{Question Writer} must provide: (1) a paper, (2) a question, (3) verbatim evidence from the source paper corresponding to each option, and (4) the correct answer. After the initial annotation, the \emph{Question Writer} evaluates the question's difficulty during a process we term the Automatic Review stage. In this stage, the writer prompts three LLMs on our platform—{\tt GPT-4o-mini}, {\tt Qwen 2.5-14B-Instruct} and {\tt GLM-4-Flash}—to answer the question based solely on the provided paper. Questions incorrectly answered by all three models proceed to Evidence Verification. Conversely, if any of the three models answers the question correctly during this stage, it is deemed insufficiently challenging and must be revised by the \emph{Question Writer} to increase its difficulty.

\subsubsection{Evidence Verification}

In this stage, \emph{Evidence Verifiers} assess whether the provided evidence excerpts are sufficient to support a correct choice. This requires: (1) analyzing and reasoning over the evidence, and (2) attempting to answer the question using only this evidence. To incentivize \emph{Question Writers} to provide accurate and logically aligned evidence, we implement a competitive mechanism: if the \emph{Evidence Verifier} arrives at the correct answer using exclusively the provided evidence, both receive Level-1 performance bonuses. Conversely, if verification fails, penalties apply: the \emph{Question Writer} and the \emph{Evidence Verifier} lose their bonus allocation. This failure triggers a \emph{Revision} cycle where the \emph{Question Writer} must improve either the evidence or the entire QA pair. The revised version must pass both Automatic Review and subsequent Evidence Verification before either party is rewarded. This filtering mechanism ensures that questions clearing this stage are \textbf{fully} answerable by human experts \textbf{given} the evidence excerpts. Crucially, this game-theoretic protocol compels \emph{Question Writers} to simultaneously elevate the difficulty standards of their questions while ensuring the provision of logically sound evidence to support their answers.

\subsubsection{Answer Verification}

In the Answer Verification phase, two constraints govern the process: \emph{correctness} and \emph{time limitation}. Specifically, the \emph{Answer Verifier} attempts to answer each question within 20 minutes using only the provided paper and without any external evidence. Based on the outcome of this process, questions are classified into three levels:
(1) \textbf{Easy}: The question is classified as easy if the verifier answers correctly within the time limit. In this case, the \emph{Question Writer} forfeits their bonus, while the \emph{Answer Verifier} receives a Level-1 bonus, indicating that the question posed an insufficient challenge.
(2) \textbf{Moderate}: The question is classified as moderate if the \emph{Answer Verifier} responds correctly but exceeds the 20-minute threshold. In this scenario, the \emph{Question Writer} receives a Level-1 bonus, while the \emph{Answer Verifier} receives no bonus, reflecting the verifier's inadequate efficiency.
(3) \textbf{Hard}: The question is classified as hard if the \emph{Answer Verifier} fails to provide the correct answer. In this case, the \emph{Question Writer} receives a Level-2 bonus, and the \emph{Answer Verifier} again receives no bonus.

In summary, the proposed mechanism induces a dynamic adversarial relationship among the three annotator groups and establishes \emph{careful annotation} as a Nash equilibrium. Specifically, any deviation from diligent annotation is identified through cross-validation against the others' annotations, consequently resulting in reduced compensation.

\subsection{Benchmark Statistics}

Tables~\ref{tab1} presents the distribution of data. For more statistics and examples, please see Appendix~\ref{datas}. 

\begin{table}[htbp]
\small
\centering
\resizebox{0.95\linewidth}{!}{%
\begin{tabular}{lcc}
    \toprule
    \textbf{Statistics}&\textbf{Numbers}&\textbf{Average Length}\\
    \midrule
    Paper&137&15,012.85\\
    Question&403&140.17\\
    \midrule
    SA-MCQ&88&163.80\\
    MA-MCQ&315&133.57\\
    \midrule
    Easy Question&85&140.05\\
    Moderate Question&109&139.82\\
    Hard Question&209&140.40\\
    \midrule
    ML Question&129&138.06\\
    CV Question&54&139.44\\
    NLP Question&220&141.59\\
    \bottomrule
\end{tabular}
}
\caption{Statistics of \textsc{ElaipBench}.}
\label{tab1}
\end{table}

\section{Experiments}

\begin{table*}[t]
\centering
\small
\resizebox{\textwidth}{!}{%
\begin{tabular}{llccccccccc}
\toprule
\multirow{2}{*}{\textbf{Paradigm}} & \multirow{2}{*}{\textbf{Models}} & \multirow{2}{*}{\textbf{Total}} & \multicolumn{2}{c}{\textbf{Question Type}} & \multicolumn{3}{c}{\textbf{Difficulty}} & \multicolumn{3}{c}{\textbf{Discipline}} \\

\cmidrule(r){4-5} \cmidrule(r){6-8} \cmidrule(r){9-11} 

 &  &  & SA-MCQ & MA-MCQ & Easy & Moderate & Hard & \multicolumn{1}{l}{ML} & \multicolumn{1}{l}{CV} & \multicolumn{1}{l}{NLP} \\ 

\cmidrule(r){1-2} \cmidrule(r){3-3} \cmidrule(r){4-5} \cmidrule(r){6-8} \cmidrule(r){9-11} 
\multirow{7}{*}{Base Models} & Llama3.3-70B-Instruct & 38.71 & 43.18 & 37.46 & 80.00 & \textbf{79.82} & 0.48 & 36.43 & \textbf{38.89} & 42.93 \\
 & Qwen3-8B & 35.48 & 23.86 & 38.73 & 69.41 & 77.06 & 0.00 & 31.01 & 35.19 & 40.98 \\
 & Qwen3-14B & 36.23 & 28.41 & 38.41 & 77.65 & 73.39 & 0.00 & 32.56 & 35.19 & 41.46 \\
 & Qwen3-32B & 37.72 & 31.82 & 39.37 & 82.35 & 75.23 & 0.48 & 33.33 & \textbf{38.89} & 42.93 \\
 & DeepSeek-V3 & \textbf{39.95} & 32.95 & \textbf{41.90} & \textbf{85.88} & 78.90 & \textbf{0.96} & \textbf{37.21} & \textbf{38.89} & \textbf{44.88} \\
 & GPT-4o-0806 & 37.47 & 22.73 & 41.59 & 80.00 & 75.23 & 0.48 & 34.11 & 37.04 & 42.44 \\
 & GPT-5 & 38.71 & 31.82 & 40.63 & \textbf{85.88} & 74.31 & \textbf{0.96} & 36.43 & \textbf{38.89} & 42.93 \\ 

\cmidrule(r){1-2} \cmidrule(r){3-3} \cmidrule(r){4-5} \cmidrule(r){6-8} \cmidrule(r){9-11} 
 
\multirow{7}{*}{CoTs} & Llama3.3-70B-Instruct + CoT & 25.81 & 25.00 & 26.03 & 80.00 & 32.11 & 0.48 & 24.81 & 24.07 & 28.78 \\
 & Qwen3-8B + CoT & 35.98 & 25.00 & 39.05 & 72.94 & 76.15 & 0.00 & 30.23 & 35.19 & 42.44 \\
 & Qwen3-14B + CoT & 31.76 & 19.32 & 35.24 & 77.65 & 56.88 & 0.00 & 31.01 & 31.48 & 34.63 \\
 & Qwen3-32B + CoT & 32.75 & 23.86 & 35.24 & 82.35 & 56.88 & 0.00 & 32.56 & 33.33 & 35.12 \\
 & DeepSeek-V3 + CoT & 29.78 & 32.95 & 28.89 & 84.71 & 42.20 & \textbf{0.96} & 27.13 & 29.63 & 33.66 \\
 & GPT-4o-0806 + CoT & 25.31 & 12.50 & 22.54 & 78.82 & 31.19 & 0.48 & 22.48 & 25.93 & 28.78 \\
 & GPT-5 + CoT & 35.24 & 22.73 & 38.73 & \textbf{85.88} & 62.39 & 0.48 & 31.78 & 35.19 & 40.00 \\ 

\cmidrule(r){1-2} \cmidrule(r){3-3} \cmidrule(r){4-5} \cmidrule(r){6-8} \cmidrule(r){9-11} 
 
\multirow{7}{*}{LRMs} & QwQ-32B & 9.68 & 12.50 & 8.89 & 31.76 & 11.01 & 0.00 & 9.30 & 9.26 & 10.73 \\
 & Qwen3-235B-A22B-thinking & 19.35 & 20.45 & 19.05 & 57.65 & 22.94 & 0.48 & 17.83 & 18.52 & 21.95 \\
 & DeepSeek-R1 & 25.81 & 31.82 & 24.13 & 61.18 & 45.87 & \textbf{0.96} & 23.26 & 25.93 & 29.27 \\
 & GPT-5-thinking-all & 37.22 & \textbf{46.59} & 34.60 & 83.53 & 70.64 & \textbf{0.96} & \textbf{37.21} & \textbf{38.89} & 39.51 \\
 & GPT-o1-mini & 35.48 & 28.41 & 37.46 & 78.82 & 67.89 & \textbf{0.96} & 34.11 & 35.19 & 39.02 \\
 & Gemini-2.5-flash-thinking & 19.35 & 18.18 & 19.68 & 60.00 & 23.85 & 0.48 & 17.05 & 20.37 & 21.95 \\
 & Claude-3.7-Sonnet-thinking & 38.46 & 43.18 & 37.14 & \textbf{85.88} & 73.39 & \textbf{0.96} & 35.66 & \textbf{38.89} & 42.93 \\ 

\cmidrule(r){1-11} 

\rowcolor[HTML]{DAE8FC} 
 
Human & - & 48.14 & 56.82 & 45.71 & 100.00 & 100.00 & 0.00/100.00* & 45.74 & 51.85 & 52.20 \\ 

\bottomrule

 &  & \multicolumn{1}{l}{} & \multicolumn{1}{l}{} & \multicolumn{1}{l}{} & \multicolumn{1}{l}{} & \multicolumn{1}{l}{} & \multicolumn{1}{l}{} &  &  & 
\end{tabular}
}
\caption{Accuracy of LLMs on the \textsc{ElaipBench}. \textbf{Bold} indicates the best result among LLMs. The \textbf{Human} results are provided by the \emph{Answer Verifier}, and values marked with * indicate the accuracy of the \emph{Evidence Verifier}.
}
\label{tab2}
\end{table*}

In this section, we evaluate the performance of frontier LLMs on \textsc{ElaipBench} and identify five key findings:
(1) Large reasoning models (LRMs) with an explicit thinking mode often underperform their base-model counterparts—a phenomenon we term \emph{reasoning paralysis} (Section~\ref{main_results}).
(2) Over half of \emph{reasoning paralysis} cases stem from harmful verification, in which models overturn initially correct answers during the reasoning process (Section~\ref{reasoning_error_analysis}).
(3) Longer reasoning chains do not necessarily improve accuracy; instead, excessively long chains often lead to incorrect results (Section~\ref{acc_length}).
(4) Models fail to adapt their reasoning depth to the difficulty of a question, producing chains of similar length regardless of its complexity (Section~\ref{difficulty_length}).
(5) Retrieval-augmented generation (RAG) yields marginal gains at best, as retrievers struggle to find relevant evidence in academic papers while models fail to sufficiently integrate the retrieved content (Section~\ref{rag}).
These findings underscore the challenges \textsc{ElaipBench} poses for academic context understanding in LLMs.

\subsection{Baseline LLMs}
We evaluate seven base models and seven LRMs. The base models include Llama3.3-70B-Instruct~\cite{dubey2024llama}, Qwen3-8B, Qwen3-14B, Qwen3-32B~\cite{yang2025qwen3}, DeepSeek-V3~\cite{liu2024deepseek}, GPT-4o-0806~\cite{hurst2024gpt}, and GPT-5\footnote{www.openai.com/index/introducing-gpt-5}. The LRMs comprise QwQ-32B~\cite{wake2024yi}, Qwen3-235B-A22B-thinking~\cite{yang2025qwen3}, DeepSeek-R1~\cite{liu2024deepseek}, GPT-5-thinking-all, GPT-o1-mini~\cite{jaech2024openai}, Gemini-2.5-flash-thinking~\cite{comanici2025gemini}, and Claude-3.7-Sonnet-thinking\footnote{www.anthropic.com/news/claude-3-7-sonnet}. To reduce variability introduced by stochastic decoding, we report the mean results from 3 independent API calls. Evaluation metric, prompts and detailed parameter configurations of all LLMs are provided in Appendix~\ref{models}.

\subsection{Results for LLMs}
\label{main_results}
We employ three primary paradigms: a base model paradigm, a chain-of-thought (CoT) paradigm, and an LRM paradigm. In the base model paradigm, models are instructed to output only the final answer option. In the CoT paradigm, we add the prompt \textit{``Let's think step by step.''} to elicit stepwise reasoning before the final answer. In the LRM paradigm, the model is allowed to produce an unstructured reasoning chain before the answer. 

As shown in Table~\ref{tab2}, the overall performance of evaluated models on \textsc{ElaipBench} is poor, with the highest accuracy reaching only 39.95\%, highlighting their limitations in handling complex academic reasoning. Moreover, integrating reasoning strategies into base models yields inconsistent improvements: apart from Qwen3-8B, all other LLMs exhibit degraded performance with CoT prompting. This trend persists in specialized LRMs—GPT-5-thinking-all underperforms GPT-5, while DeepSeek-R1 trails DeepSeek-V3. This phenomenon, which we term \emph{reasoning paralysis}, indicates that current reasoning paradigms are insufficient for \textsc{ElaipBench}'s academic challenges.

In both SA-MCQ and MA-MCQ tasks, LLMs consistently underperform human experts. In SA-MCQ, the LRMs demonstrate moderate improvement over the base model (GPT-5-thinking-all achieves 46.59\%, surpassing GPT-5’s 31.82\%).  However, in the MA-MCQ setting, the accuracy of LLMs drops significantly. We hypothesize that this decline is attributable to the increased complexity of the MA-MCQ format, which, by requiring the selection of multiple correct answers, makes it more challenging for LLMs to reliably identify all correct options. Furthermore, no LLM exceeds human expert performance on questions of easy or moderate difficulty. On hard questions, the marginal advantages exhibited by a few models (a mere 0.96\%) are insufficient to warrant their reliable use by human experts for academic paper understanding. Finally, the consistent underperformance of LLMs compared to human experts across all AI-related disciplines indicates that their limitations in paper comprehension are generalizable.

\begin{figure}[t]
  \centering
  \includegraphics[width=0.9\linewidth]{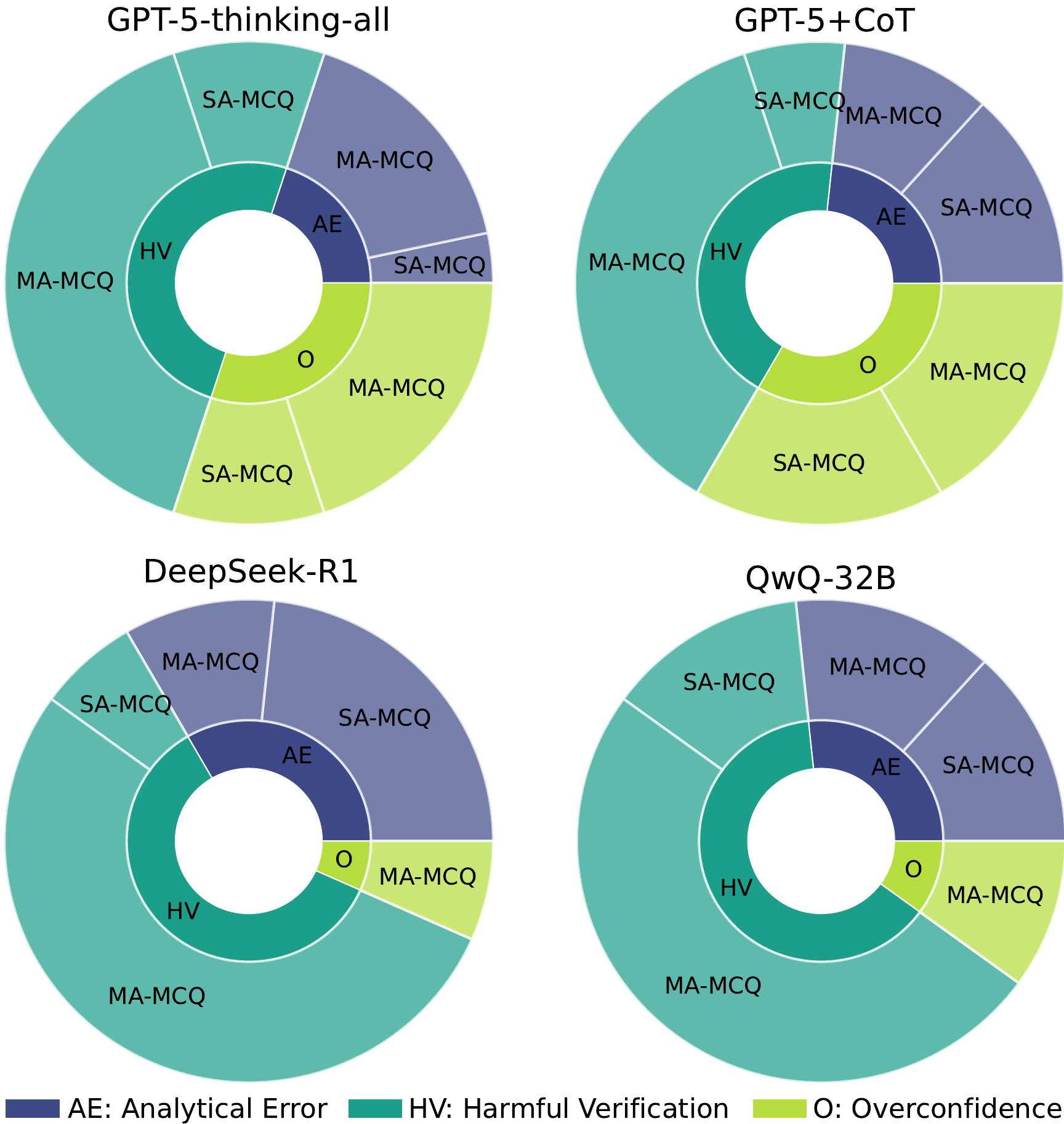}
  \caption{Reasoning error distribution of \textsc{ElaipBnech} on different LLMs.}
  \label{fig3}
\end{figure}

\subsection{Reasoning Error Analysis}
\label{reasoning_error_analysis}

To investigate the causes of \emph{reasoning paralysis}, we randomly selected 30 incorrectly answered questions from the results produced under both the LRM and CoT paradigms and conducted a manual examination of the outputs. Based on our analysis of these errors, we classify the causes into three categories: (1) \textbf{Analytical Error}, where the model consistently performs incorrect reasoning to produce wrong answers; (2) \textbf{Harmful Verification}, where the model's flawed corrective measures during its verification process alter originally correct answers to incorrect ones; and (3) \textbf{Overconfidence}, where the model determines answers based on its own prior knowledge rather than the provided text. Figure~\ref{fig3} illustrates the distribution of these failure types. Notably, across all integrated reasoning settings, \textbf{Harmful Verification} accounts for over half of all error cases, indicating that the model often initially generates the correct answer but subsequently invalidates it through excessive re-analysis. This finding suggests that \textbf{Harmful Verification} is the primary reason for the underperformance of the reasoning paradigms compared to the base models. Furthermore, models exhibit a higher tendency toward \emph{reasoning paralysis} in MA-MCQ tasks. Detailed examples of these three error types are provided in Appendix~\ref{error_analysis}.

\subsection{Accuracy vs. Reasoning Length} 
\label{acc_length}

\begin{figure}[htbp]
  \centering
  \includegraphics[width=0.9\linewidth]{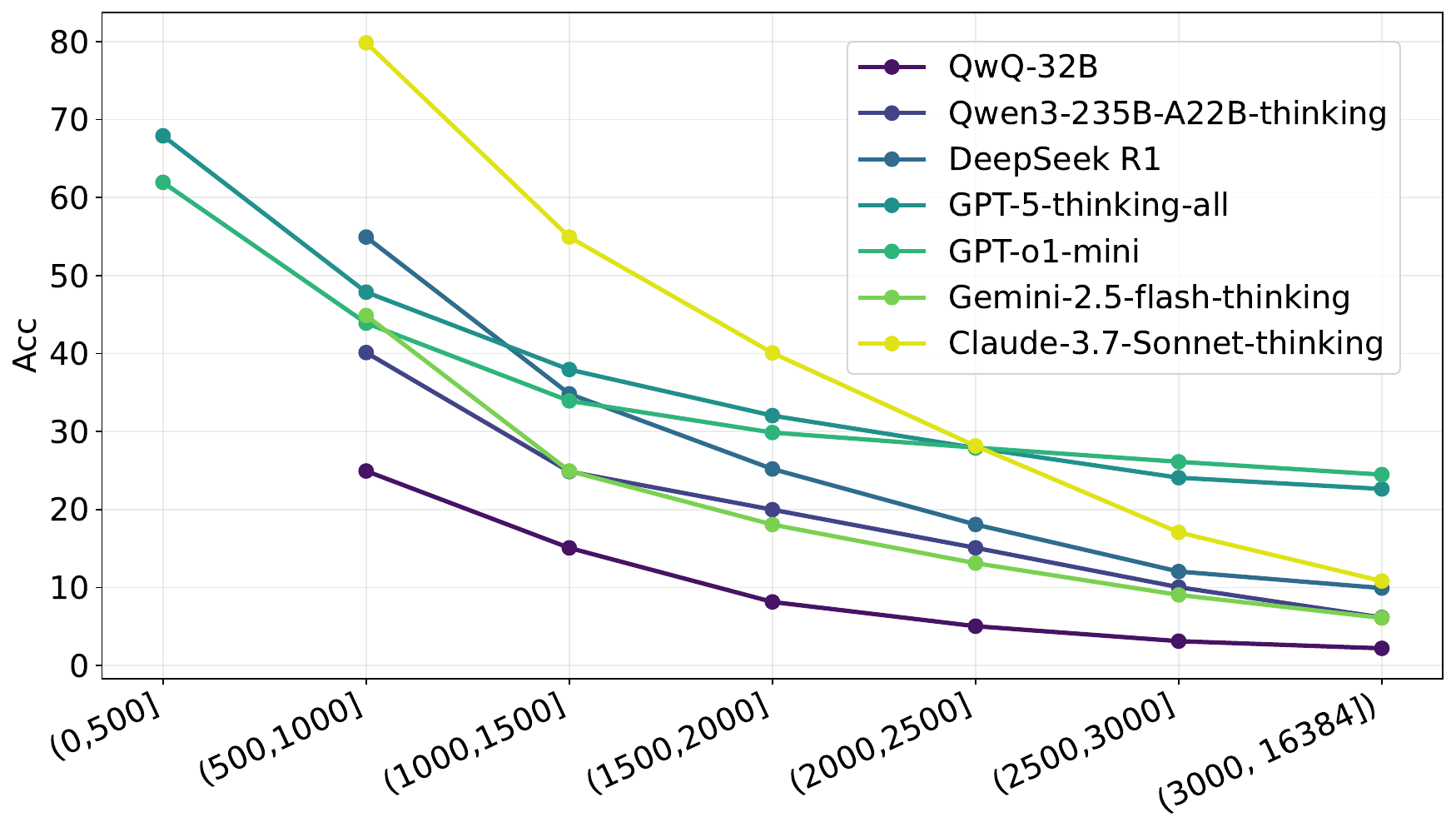}
  \caption{Completion Tokens vs.Performance.}
  \label{fig4}
\end{figure}

Figure~\ref{fig4} shows the relationship between reasoning length and accuracy across different LRMs. A consistent trend emerges: as the output length increases from shorter to longer ranges, accuracy does not improve for any model; instead, it steadily declines. This degradation is most pronounced for Claude-3.7-Sonnet-thinking. These results at least indicate that longer outputs are not a reliable indicator of more thorough reasoning; rather, they often reflect a tendency to generate additional explanations, branching hypotheses, or redundant steps when the model is uncertain. We further observe that self-verification behaviors account for a substantial fraction of LRM reasoning traces (approximately 50\%). Such segments often begin with phrases such as \emph{wait} or \emph{Alternatively}, followed by a restatement of the original question and a restart of the reasoning process. This repetitive pattern substantially lengthens the reasoning path while introducing largely ineffective verification steps. Representative cases are provided in Appendix~\ref{error_analysis}, together with statistics summarizing the average reasoning length across models.

\subsection{Difficulty vs. Reasoning Length}
\label{difficulty_length}

\begin{figure}[htbp]
  \centering
  \includegraphics[width=0.9\linewidth]{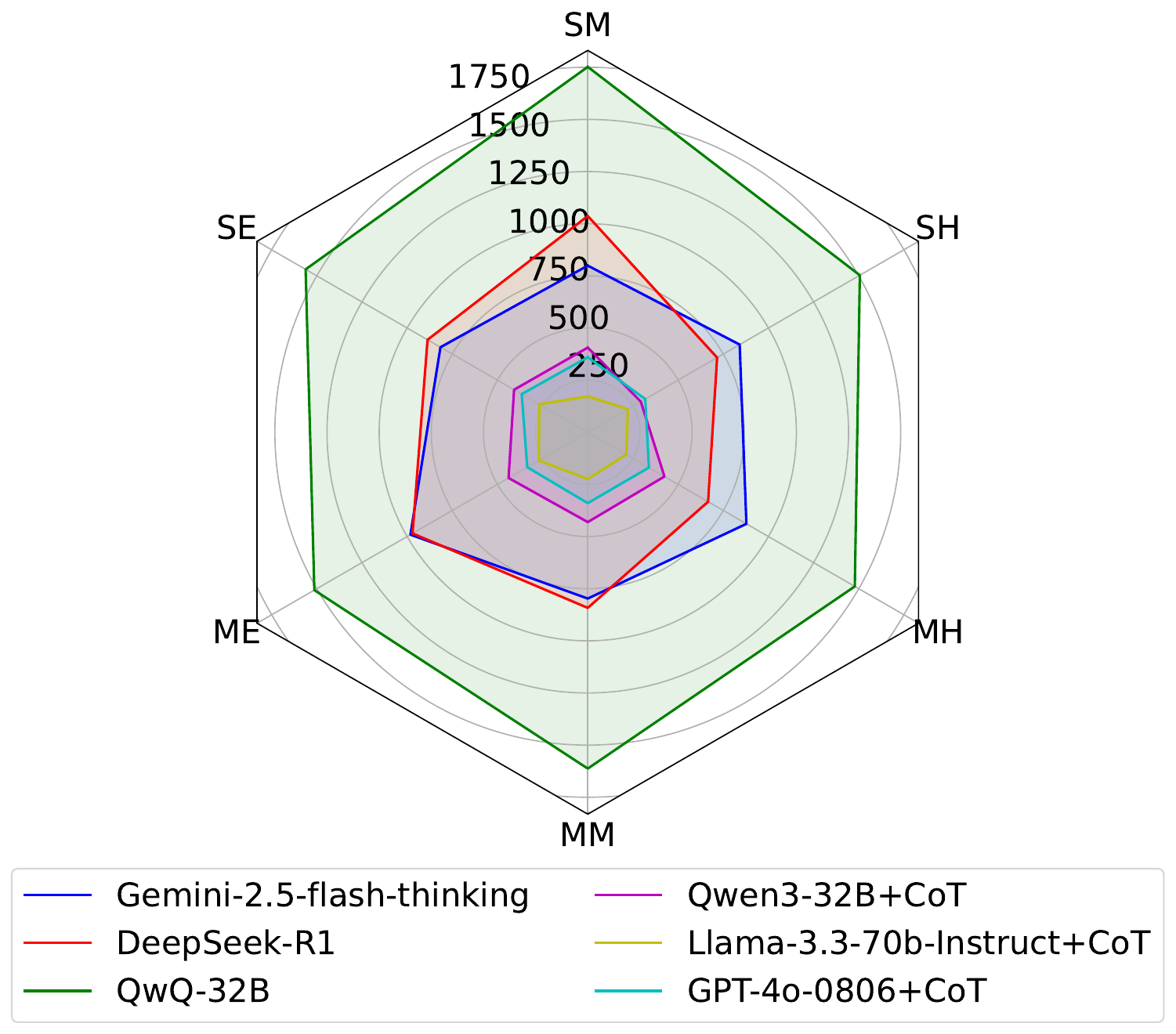}
  \caption{The reasoning lengths across different question types (SE: SA-MCQ + easy, SM: SA-MCQ + moderate, SH: SA-MCQ + hard, ME: MA-MCQ + easy, MM: MA-MCQ + moderate, MH: MA-MCQ + hard).}
  \label{fig5}
\end{figure}

Figure~\ref{fig5} presents the reasoning lengths of LLMs across different question types. We observe that all LRMs generate longer outputs than the CoT paradigm. However, across question categories—including SA-MCQ, MA-MCQ, easy, moderate, and hard difficulty levels—none of the baselines exhibit meaningful variation in reasoning length. This lack of adaptive behavior suggests that current LLMs fail to recognize the relationship between distractors in the options and the paper content, and consequently do not engage in deeper analysis. This shallow reasoning process often leads to incorrect inferences, which in turn degrades final performance, resulting in reasoning-enhanced LLMs sometimes underperforming non-reasoning baselines. These findings underscore a critical limitation in current LLMs: while they can produce longer responses, they lack the ability to strategically adjust reasoning depth based on question complexity.

\subsection{RAG Performance}
\label{rag}

In our RAG-based evaluation, we adopt two retrieval paradigms: intra-paper retrieval and web-based retrieval. For intra-paper retrieval, we segment each paper into non-overlapping passages of 512 tokens using sentence boundaries as delimiters. We then retrieve the top-five most relevant passages for each question using two methods: dense passage retrieval (DPR) based on the {\tt BGE-m3}~\cite{chen2024m3} encoder and the BM25 algorithm~\cite{robertson2009probabilistic}. For web-based retrieval, we query the Google API using the question and paper title as input, retaining the content of the top-five most relevant web pages. In the prompt, we explicitly inform the LLM that the retrieved content constitutes auxiliary knowledge, which it may consult alongside the original paper. Detailed hyper-parameters, RAG configurations and prompts are provided in Appendix~\ref{rag_settings}.

Figure~\ref{fig6} presents the impact of RAG on answer accuracy across different LLMs. We observe that both {\tt BGE-m3}-based DPR and BM25 intra-paper retrieval lead to performance degradation. This stems from the retrievers’ inability to effectively align the questions with paper content, making it difficult to pinpoint evidence strongly correlated with the answer options.
Within the web-based paradigm, improvements in answer accuracy remain marginal. Our analysis of the retrieved results reveals that nearly all returned documents are other relevant academic papers, compelling the LLM to perform multi-document QA while simultaneously analyzing complex scholarly content. Moreover, many models struggle to extract and synthesize information from noisy web sources, impeding coherent and contextually grounded reasoning. Although models such as GPT-5, GPT-5-thinking-all, and Claude-3.7-Sonnet-thinking show modest performance gains, these incremental improvements fall short of the expert-level academic understanding expected.
We believe that a carefully engineered retriever, coupled with a structured organization of retrieved external knowledge, could enhance the capacity of LLMs for academic paper comprehension. Nevertheless, optimizing RAG system configurations lies beyond the scope of our study, and thus we defer such investigations to future work.

\begin{figure}[t]
  \centering
  \includegraphics[width=0.9\linewidth]{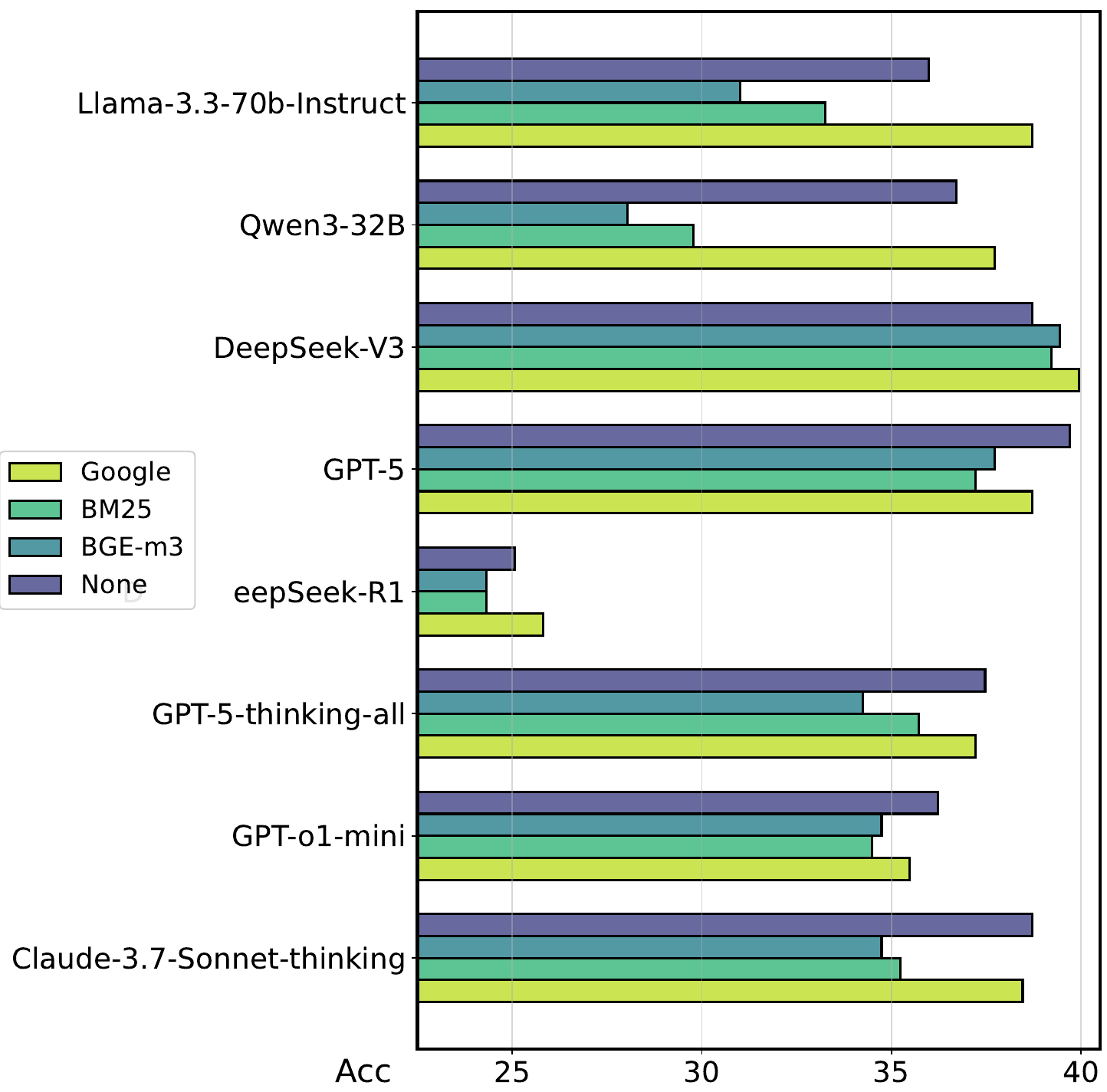}
  \caption{RAG Performance. The label \textbf{None} indicates that the LLM generates answers directly, and its accuracy corresponds to the \textbf{Total} reported in Table~\ref{tab2}.}
  \label{fig6}
\end{figure}

\section{Conclusion}
In this work,  we introduce \textsc{ElaipBench}, an expert-level benchmark designed to evaluate the understanding of AI research papers by LLMs. Our evaluation reveals that even state-of-the-art LLMs achieve only moderate performance (best at 39.95\%), highlighting a substantial gap in current models’ ability to perform fine-grained, evidence-grounded reasoning over complex scientific texts.
\section{Limitations}

We acknowledge several limitations of our work. (1) The scope of our benchmark is constrained in two ways: it is limited to the AI domain and covers only a single modality—text. However, expanding the benchmark is hindered by the practical challenges and high cost of annotation; for instance, recruiting expert annotators to collect the current 403 questions required six months and over 50,000 CNY (\textasciitilde7,000 USD). (2) The benchmark is English-only, which limits the evaluation of models' cross-lingual capabilities. In the future, we aim to expand the benchmark in two directions: including other scientific disciplines such as biology, physics, and chemistry, and incorporating multi-modality to enable a broader evaluation of scientific comprehension.

\bibliography{ref}

\clearpage
\appendix
\section{Appendix}

\subsection{Academic Paper Understanding Dataset Comparison}
\label{dataset_comparsion}

As shown in Table~\ref{tab:academic_datasets}, we summarize key dataset characteristics along multiple dimensions, including whether the benchmark is constructed from full-length papers, the number of instances, task format, the presence of objective evaluation metrics, and whether the data are human-annotated.

\begin{table*}[htbp]
\centering
\resizebox{\textwidth}{!}{%
\rowcolors{2}{gray!12}{white}
\begin{tabular}{lccccc}
\toprule
\textbf{Datasets} & \textbf{Full-Length Paper} & \textbf{\#Data} & \textbf{Task} & \textbf{Objective Evaluation Metric} & \textbf{Human Annotation} \\
\midrule
ArxivBench~\cite{li2025arxivbench}& \xmark & 6,500 & Open QA & \xmark & \xmark \\
SciQAG~\cite{wan2024sciqag}& \xmark & 188,042 & Open QA & \xmark & \xmark \\
ChemLit-QA~\cite{wellawatte2025chemlit} & \xmark & 1,000 & Open QA & \xmark & \xmark \\
ScienceQA~\cite{saikh2022scienceqa} & \xmark & 100k & Open QA & \xmark & \xmark \\
LITQA~\cite{lala2023paperqa} & \xmark & 50  & MCQ & \cmark & \cmark \\
PaperBench~\cite{starace2025paperbench} & \cmark & 20 & Code Generation & \cmark & \cmark \\
AcadReason~\cite{gui2025acadreason} & \cmark & 50 & Open QA & \xmark & \cmark \\
\midrule
ELAIPBench   & \cmark & 403     & MCQ             & \cmark & \cmark \\ 
\bottomrule
\end{tabular}
}
\caption{Academic paper understanding dataset.}
\label{tab:academic_datasets}
\end{table*}

Among the listed benchmarks, only a small subset—PaperBench~\cite{starace2025paperbench}, AcadReason~\cite{gui2025acadreason}, and our proposed ELAIPBench—is built on complete scientific papers. This distinction is crucial: understanding an entire paper requires sustained comprehension of complex methodology, nuanced argumentation, and long-range dependencies, which cannot be adequately assessed using only abstracts, snippets, or short prompts. By contrast, ArxivBench~\cite{li2025arxivbench}, SciQAG~\cite{wan2024sciqag}, ChemLit-QA~\cite{wellawatte2025chemlit}, and ScienceQA~\cite{saikh2022scienceqa} rely on short or automatically generated inputs that may measure retrieval or surface-level knowledge, but are insufficient for evaluating deep understanding over full academic texts.

Dataset scale further differentiates existing resources. While some benchmarks (e.g., SciQAG) contain more than 180,000 instances, these are automatically generated and often lack the authenticity and depth required to assess expert-level reasoning. In contrast, human-annotated datasets are typically much smaller due to high cost and domain expertise requirements. For example, LITQA~\cite{lala2023paperqa} contains only 50 MCQs and does not provide full-paper context; PaperBench includes only 20 instances and primarily targets code generation rather than comprehension. Although PaperBench places substantial demands on planning, engineering, and coding ability, these skills are largely orthogonal to the central goal of evaluating scientific understanding. AcadReason is based on full papers but likewise includes only 50 questions, limiting its statistical power for robust model evaluation.

Task design and evaluability also vary substantially across benchmarks. Many datasets, including ArxivBench, SciQAG, and AcadReason, adopt open QA settings, making objective and reproducible evaluation difficult. Accordingly, we approach such paradigms with caution when aiming for rigorous benchmarking.

Finally, human annotation remains essential for producing high-quality, domain-specific questions grounded in genuine scientific reasoning. Although automatic generation can scale efficiently, it often falls short in the subtlety and cognitive rigor needed for expert assessment. Our annotation protocol further strengthens reliability via a game-theoretic mechanism that incentivizes careful reading and verification, thereby excluding superficial engagement that can arise in pipeline-style labeling.

In summary, ELAIPBench uniquely synthesizes four key attributes: (1) the utilization of full-text papers, (2) large-scale human annotation, (3) objective evaluation via MCQs, and (4) a focus on deep comprehension rather than citation retrieval or code generation. These properties make ELAIPBench a rigorous and reliable tool for evaluating advanced LLMs in realistic academic comprehension scenarios.

\subsection{Annotation Guidelines}
\label{guidelines}
All annotators utilized the annotation platform anonymously, ensuring their identities remained concealed from one another and thereby eliminating the possibility of collusion for reward acquisition. Different annotators are provided with distinct guidelines based on their assigned tasks. The content of these guidelines is shown below:

\begin{tcolorbox}[
    colback=gray!10,
    colframe=darkgray,
    title=Guideline for Question Writer,
    coltitle=white,
    fonttitle=\bfseries,
    boxrule=1.5pt,
    width=\columnwidth,
    before skip=1em,
    after skip=1em,
    breakable
]

\paragraph{Overview}  
Thank you for participating in this annotation project. Your task is to read the assigned academic paper carefully and write well-designed, challenging questions based on its content. These questions will be used to evaluate the academic reasoning capabilities of large language models (LLMs) in long-context settings.
Your contribution is highly valuable to the advancement of AI benchmarks. We ask you to approach this task with care and precision, as your work directly supports the development of more accurate and responsible AI systems.

\paragraph{General Guidelines}
\begin{itemize}
  \item Prefer documents between 8,192 and 128,000 words.
  \item Read the paper thoroughly before writing questions.
  \item Questions must not be answerable by simple keyword matching or shallow retrieval.
\end{itemize}

\paragraph{Question Design Requirements}
\begin{itemize}
\item Avoid selecting overly well-known or commonly cited academic papers to ensure diversity and reduce the risk of models relying on memorized knowledge. 
\item Questions must be objectively formulated in English, free of subjectivity or ambiguity.
\item The correct answer must not be obtainable through simple keyword matching or direct retrieval from the text; instead, answering should require reasoning, summarization, or integration of multiple pieces of information across the paper.
\item To prevent confusion or unintended ambiguity, questions and options must not introduce domain-specific terms or abbreviations unless they are explicitly defined within the source paper.
\item Distractor selections should be carefully designed to include misleading phrasing or partial truths—statements that align with some aspects of the paper but are ultimately incorrect—to increase cognitive demand and necessitate careful comparison and deep reasoning.
\item For SA-MCQ, exactly one option must be correct.
\item For MA-MCQ, either two or three options must be correct, ensuring a balanced level of difficulty and discouraging random guessing.
\end{itemize}

\paragraph{Reward Rules}  
Annotators will receive \textbf{30 CNY} for each well-written and verified question. However, the reward may be reduced under the following conditions:
\begin{itemize}
  \item If the question fails to pass evidence verification.
  \item If the question is too easy and can be answered correctly without reasoning.
  \item If the question or options violate any of the design guidelines above.
\end{itemize}

If your question fails the evidence verification stage, you will be required to revise and resubmit it based on the verifier’s feedback. In such cases, your reward will be reduced accordingly.

\paragraph{Final Note}  
We sincerely appreciate your effort and participation in this project. Your thoughtful contributions are essential to building more capable and trustworthy AI systems. Thank you for your cooperation!

\end{tcolorbox}

\begin{tcolorbox}[
    colback=gray!10,
    colframe=darkgray,
    title=Guideline for Evidence Verifier,
    coltitle=white,
    fonttitle=\bfseries,
    boxrule=1.5pt,
    width=\columnwidth,
    before skip=1em,
    after skip=1em,
    breakable
]

\paragraph{Overview}  
Thank you for participating in the evidence verification process. Your task is to carefully read the provided academic paper, the corresponding question, and the proposed evidence. Based on these inputs, you must attempt to answer the question as accurately as possible. This process is designed to evaluate whether the evidence is sufficient and relevant to support the question.

\paragraph{General Workflow}
\begin{itemize}
  \item Carefully read the assigned paper segment, the question, and the associated evidence.
  \item Try to answer the question solely using the provided paper content and evidence.
  \item Submit your answer through the platform interface within the given time limit.
\end{itemize}

\paragraph{Answer Judgment and Feedback}
\begin{itemize}
  \item If your answer is correct, you will receive a reward for this verification task.
  \item If your answer is incorrect, a penalty will be applied by deducting part of the reward.
  \item Incorrect answers indicate that the evidence may not sufficiently support the question.
  \item In such cases, the corresponding question will be sent back to the original question author for revision.
\end{itemize}

\paragraph{Guidelines for Verification}
\begin{itemize}
  \item Base your judgment strictly on the provided evidence and the content of the paper.
  \item Do not incorporate any external knowledge or assumptions.
  \item Make sure to complete the task within the time limit.
  \item Be precise in your answer; partial correctness may still lead to rejection.
\end{itemize}

\paragraph{Reward Rules}
\begin{itemize}
  \item You will receive a base reward for each correctly verified question.
  \item If your answer is incorrect:
    \begin{itemize}
      \item A partial deduction will be made from your total reward for this task.
      \item The question will be marked as needing revision and returned to the question annotator.
    \end{itemize}
\end{itemize}

\paragraph{Final Note}
Your participation is essential to ensuring the quality and reliability of our benchmark. Accurate evidence verification directly enhances the overall dataset quality and model evaluation fairness. We appreciate your careful attention and thoughtful work.

\end{tcolorbox}

\begin{tcolorbox}[
    colback=gray!10,
    colframe=darkgray,
    title=Guideline for Answer Verifier,
    coltitle=white,
    fonttitle=\bfseries,
    boxrule=1.5pt,
    width=\columnwidth,
    before skip=1em,
    after skip=1em,
    breakable
]

\paragraph{Overview}  
Thank you for participating in the answer verification process. Your task is to answer a question based on a given academic paper, simulating the perspective of a well-informed reader. Your answer quality and response time will be used to assess the difficulty level of the question and to determine the reward distribution for both you and the original question author.

\paragraph{General Workflow}
\begin{itemize}
  \item Read the provided paper and the associated question carefully.
  \item Answer the question using only the paper content; external knowledge is not allowed.
  \item Submit your answer within the interface, where a timer will be running.
\end{itemize}

\paragraph{Timing and Reward Mechanism}
\begin{itemize}
  \item A timer starts when the question is displayed.
  \item If you submit a correct answer within the expected time window, you will receive the maximum reward.
  \item If you exceed the time limit, a partial penalty will be applied.
  \item If your answer is incorrect, a more significant penalty will be imposed.
\end{itemize}

\paragraph{Impact on Question Difficulty}
\begin{itemize}
  \item Your performance is used to estimate the difficulty of the question:
    \begin{itemize}
      \item Quick and accurate answers suggest the question is easy.
      \item Slow or incorrect responses indicate higher question difficulty.
    \end{itemize}
  \item These estimates will influence quality control, benchmarking statistics, and bonus calculations for the original question writer.
\end{itemize}

\paragraph{Answering Guidelines}
\begin{itemize}
  \item Base your answer strictly on the information provided in the paper.
  \item Avoid speculative reasoning or assumptions beyond the document.
  \item Be concise but precise—ambiguous answers may be penalized.
\end{itemize}

\paragraph{Final Note}
Your role is crucial in validating the reliability and effectiveness of each question. Accurate and timely answers not only help refine the question set but also contribute directly to fairer model evaluation benchmarks. We appreciate your rigorous participation and attention to detail.

\end{tcolorbox}

\subsection{Annotation Details}
\label{annotation_details}

\subsubsection{Annotator Statistics}
\label{annotator_statistics}

\begin{figure}[htbp]
  \centering
  \includegraphics[width=0.9\linewidth]{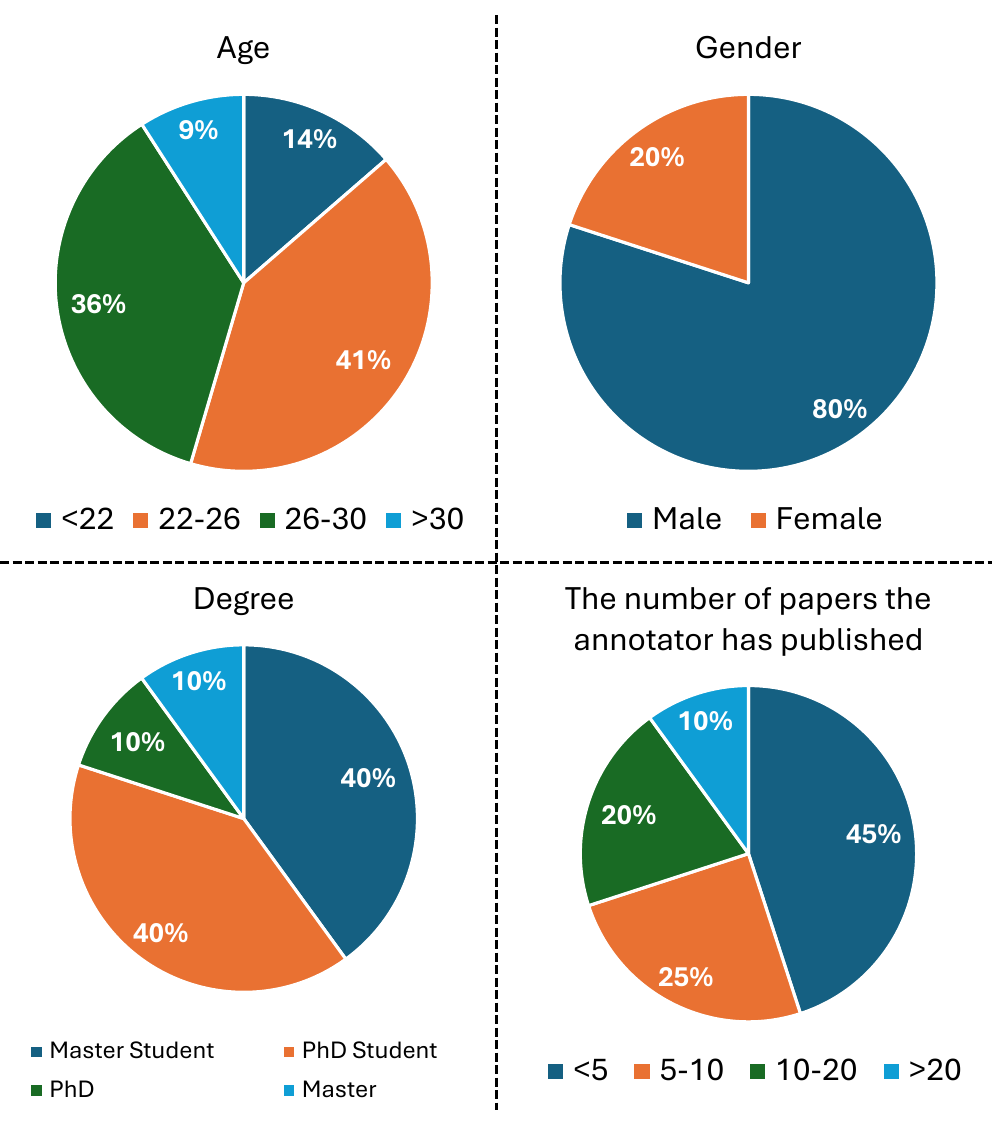}
  \caption{Distribution of our annotators across ages, genders, and academic status.}
  \label{fig7}
\end{figure}

To better understand the background and composition of our annotators, we collected information on their age, gender, and academic status during registration. Figure~\ref{fig7} summarizes the distribution across these three dimensions. In terms of age, the majority of annotators fall within the 22–26 (41\%) and 26–30 (36\%) age groups, while those under 22 (14\%) and over 30 (9\%) constitute smaller proportions. Regarding gender, 80\% of annotators identify as male and 20\% as female. 
All annotators hold or are currently pursuing a master’s degree or higher: specifically, the cohort includes 8 master’s students, 2 individuals with completed master’s degrees, 8 PhD students, 1 postdoctoral researcher, and 1 professor. Furthermore, every annotator has authored at least one academic paper, and 55\% have published more than five papers, underscoring the high level of academic expertise within the annotation team.

\subsubsection{Annotation Platform}

\begin{figure*}[htbp]
  \centering
  \includegraphics[width=\textwidth]{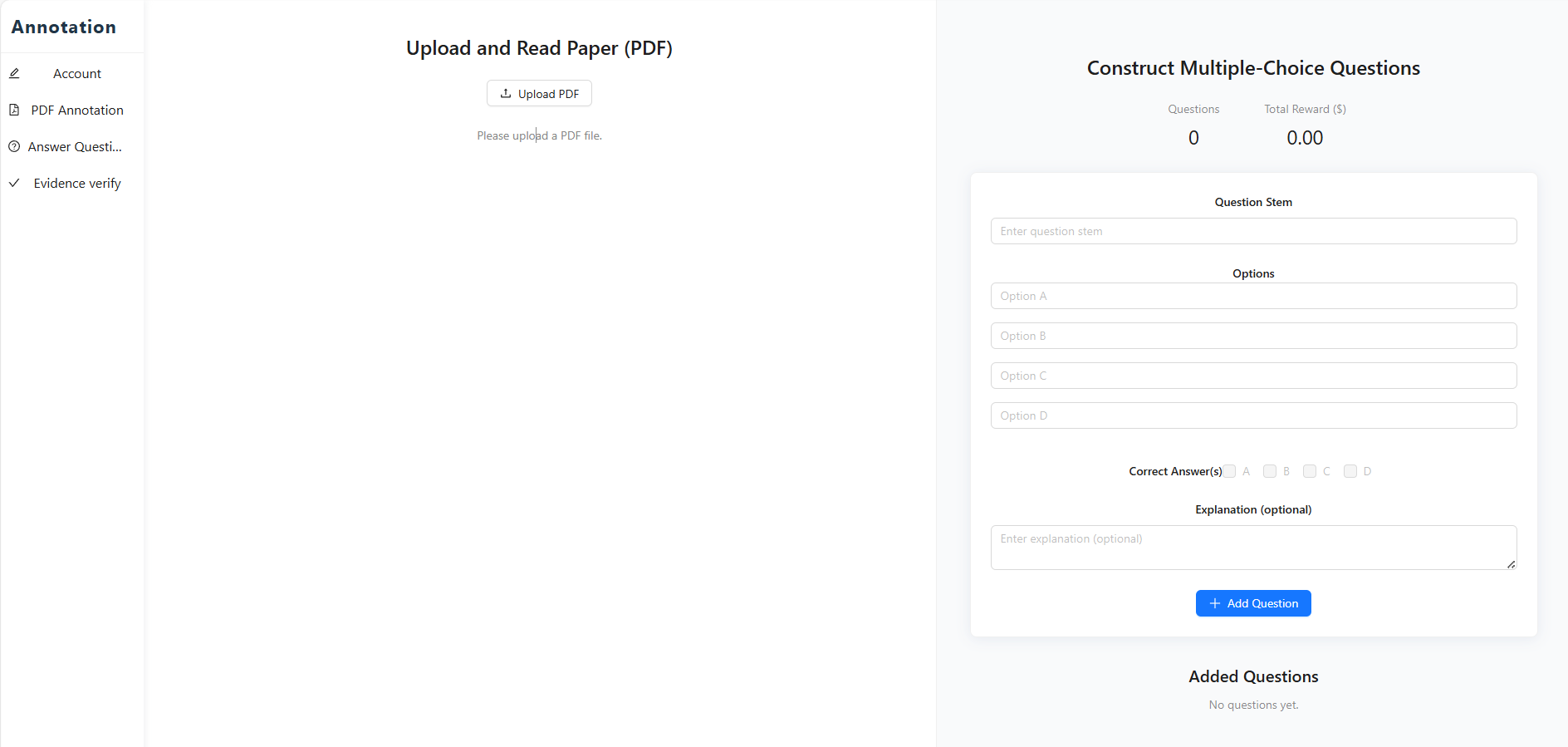}  
  \caption{Screenshot of the QA pair annotation page. After logging in, annotators can upload papers and perform annotations on this page. They can also see the number of questions they have annotated and the amount of money they have earned.}
  \label{fig8}  
\end{figure*}

\begin{figure*}[htbp]
  \centering
  \includegraphics[width=1\textwidth]{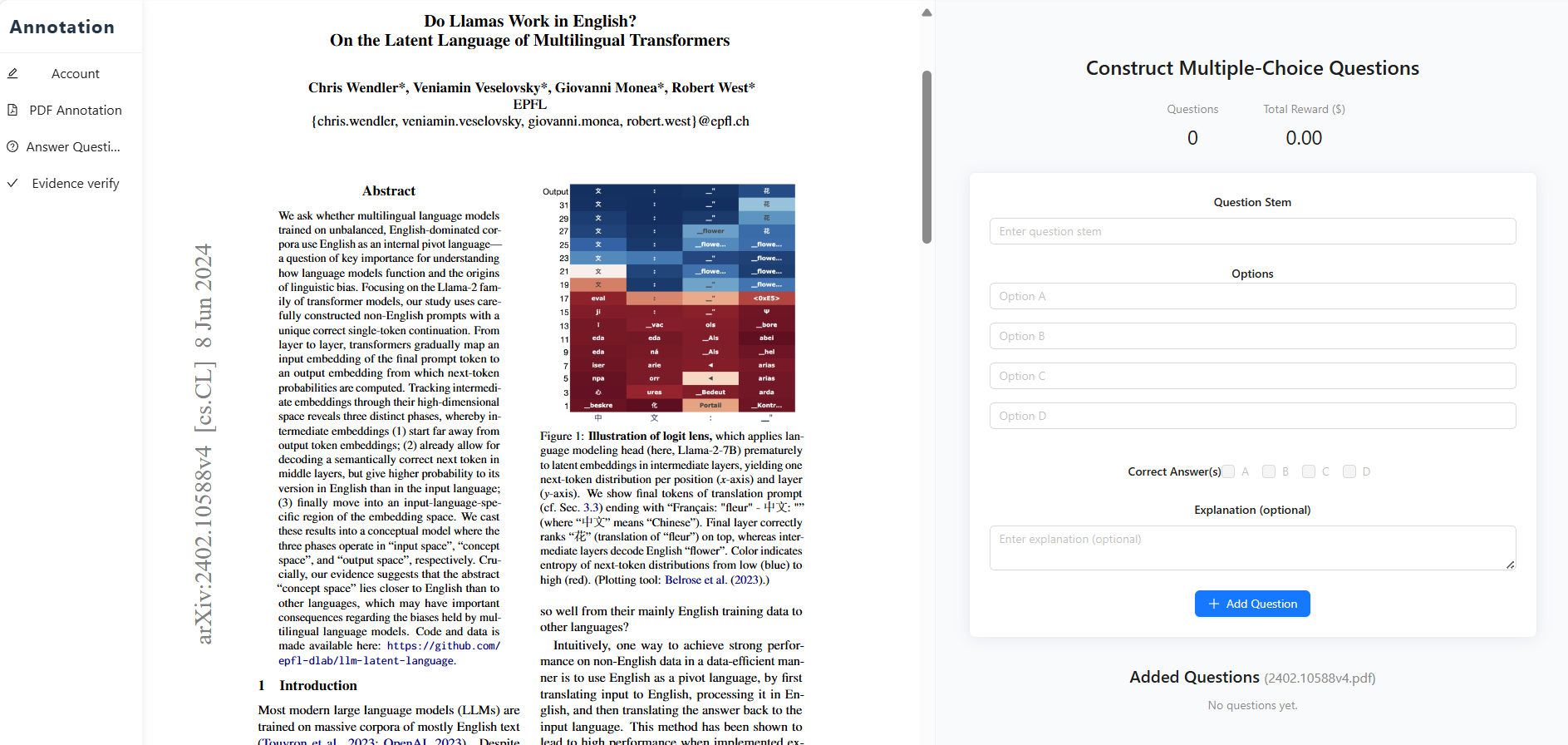}  
  \caption{Screenshot of the QA pair annotation page. After uploading the papers, \emph{Question Writers} can browse the document and create QA pairs along with corresponding evidence on the right side of the page.}
  \label{fig9}  
\end{figure*}

\begin{figure*}[htbp]
  \centering
  \includegraphics[width=\textwidth]{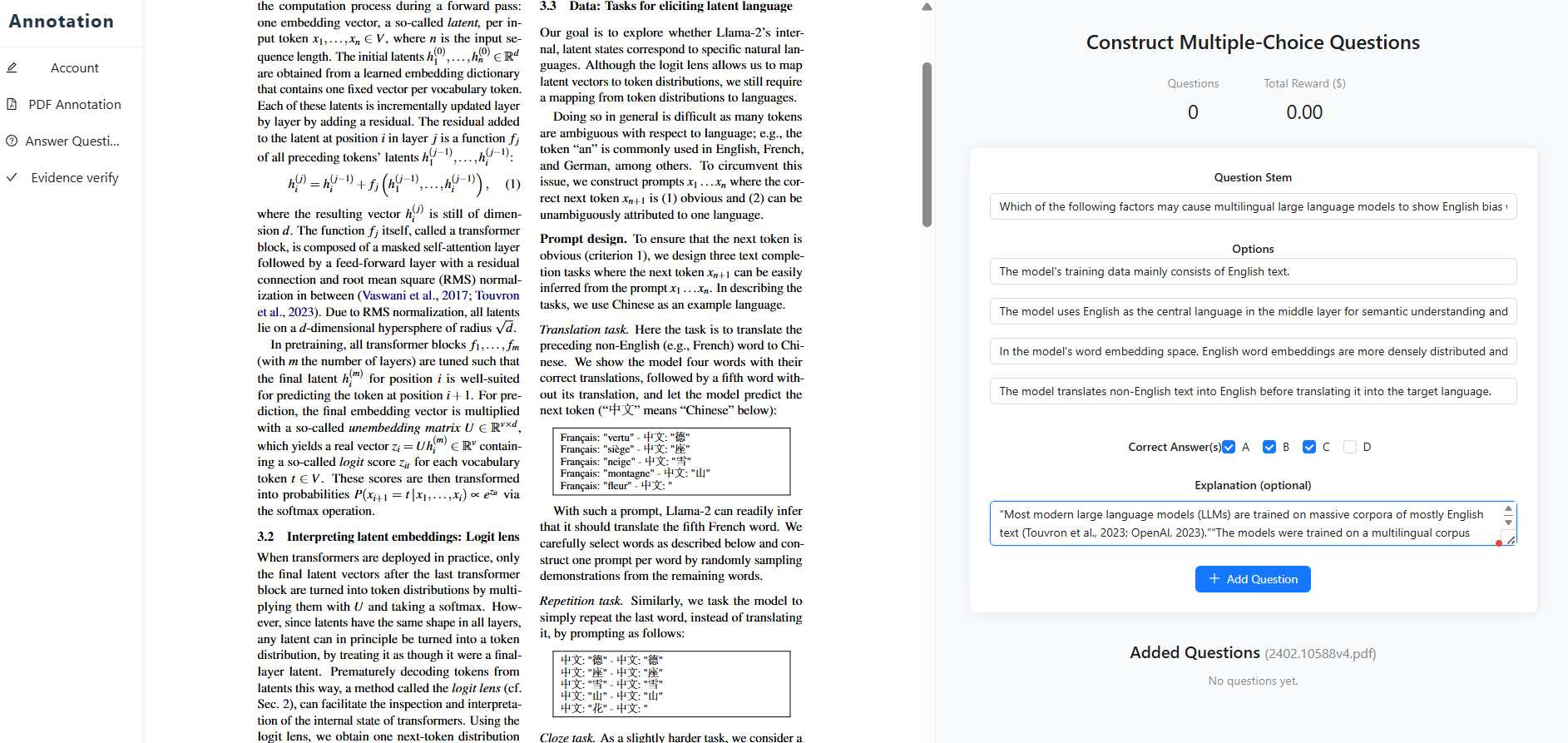}  
  \caption{Screenshot of the QA pair annotation page. Annotators input the created QA pairs.}
  \label{fig10}
\end{figure*}

\begin{figure*}[htbp]
  \centering
  \includegraphics[width=\textwidth]{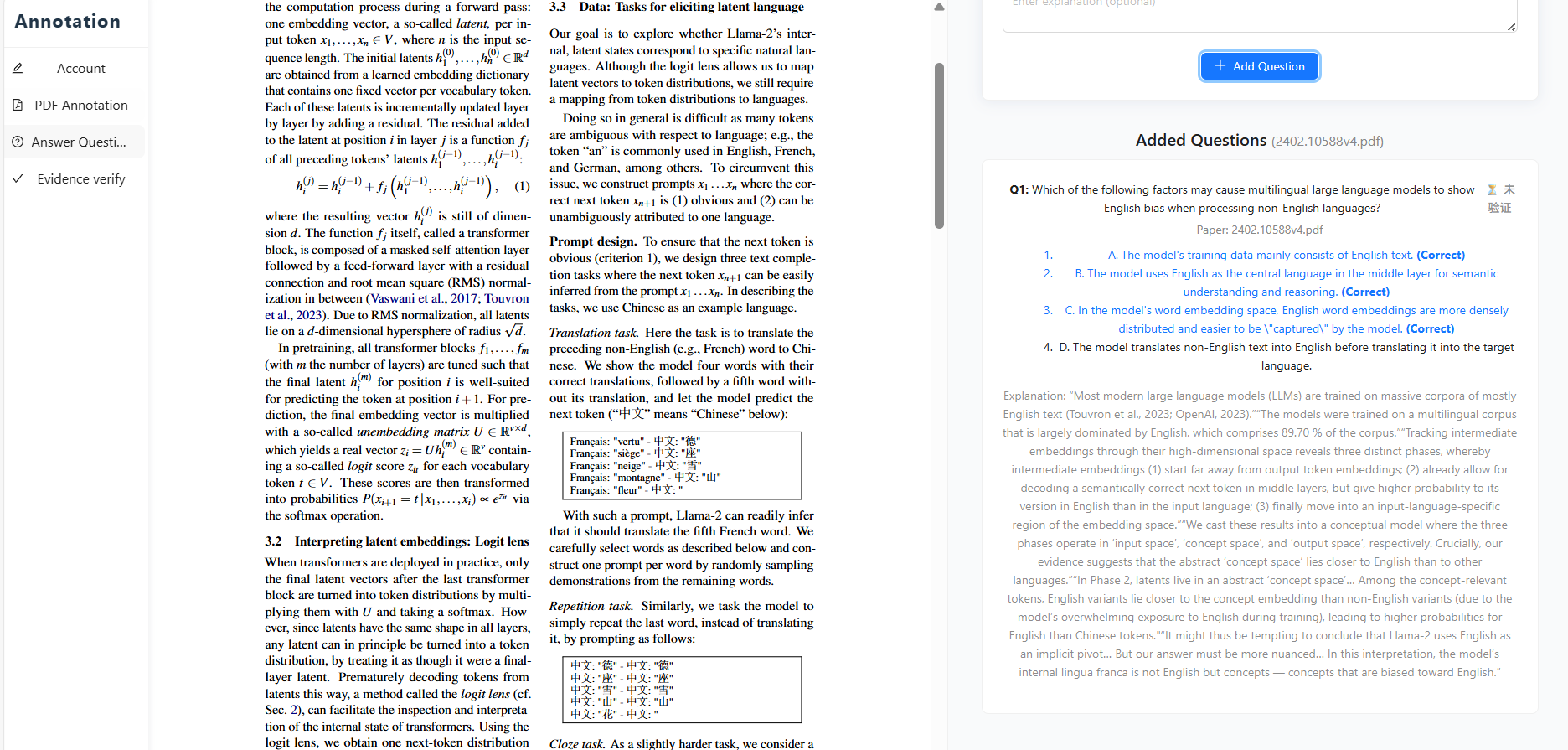}  
  \caption{Screenshot of the QA pair annotation page. After annotating and submitting questions, annotators can view the questions they have submitted and the amount of money they have earned.}
  \label{fig11}
\end{figure*}

\begin{figure*}[htbp]
  \centering
  \includegraphics[width=\textwidth]{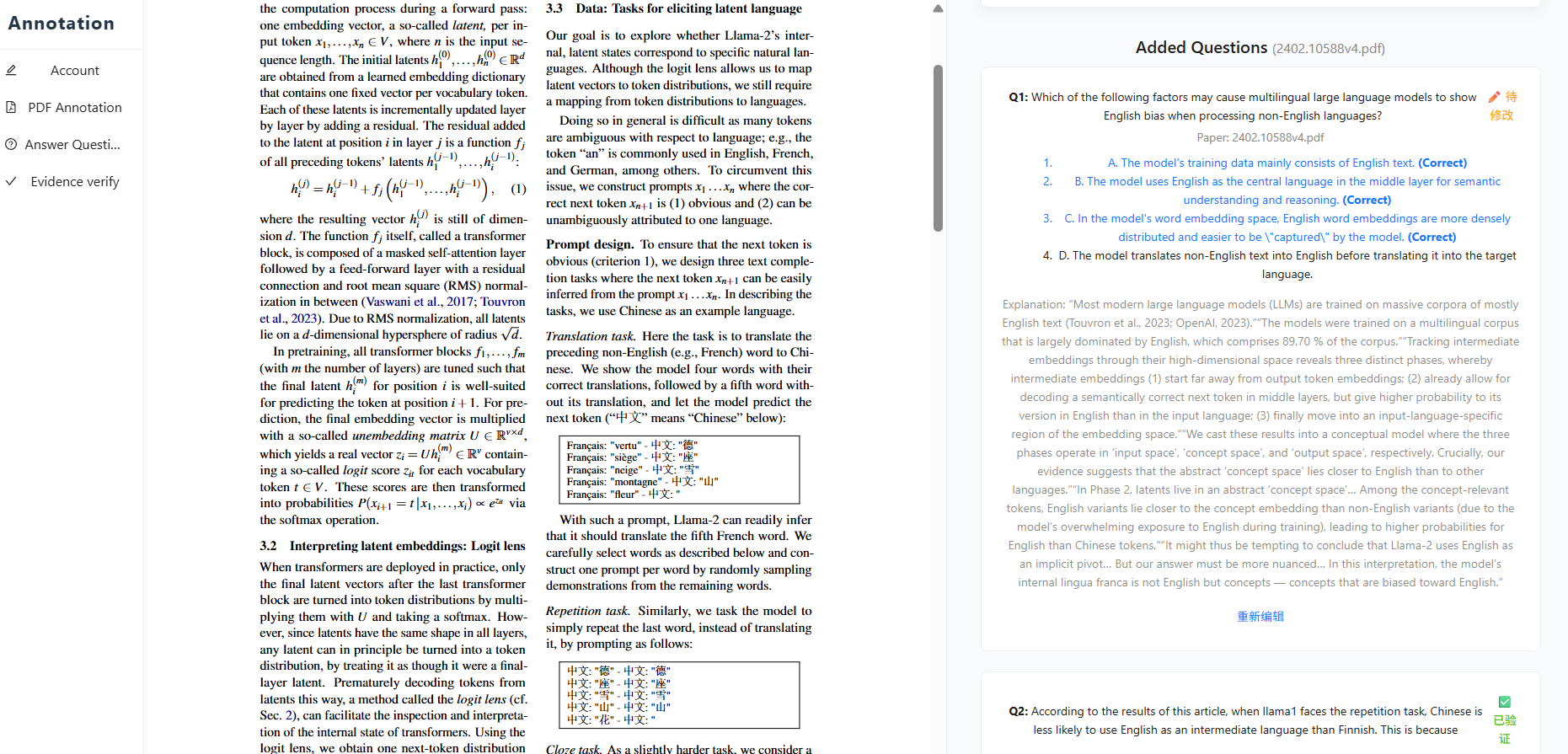}  
  \caption{Screenshot of the QA pair annotation page. After verification of evidence, annotators can check the status of the questions submitted and revise those that did not pass the verification.}
  \label{fig12}
\end{figure*}

\begin{figure*}[htbp]
  \centering
  \includegraphics[width=\textwidth]{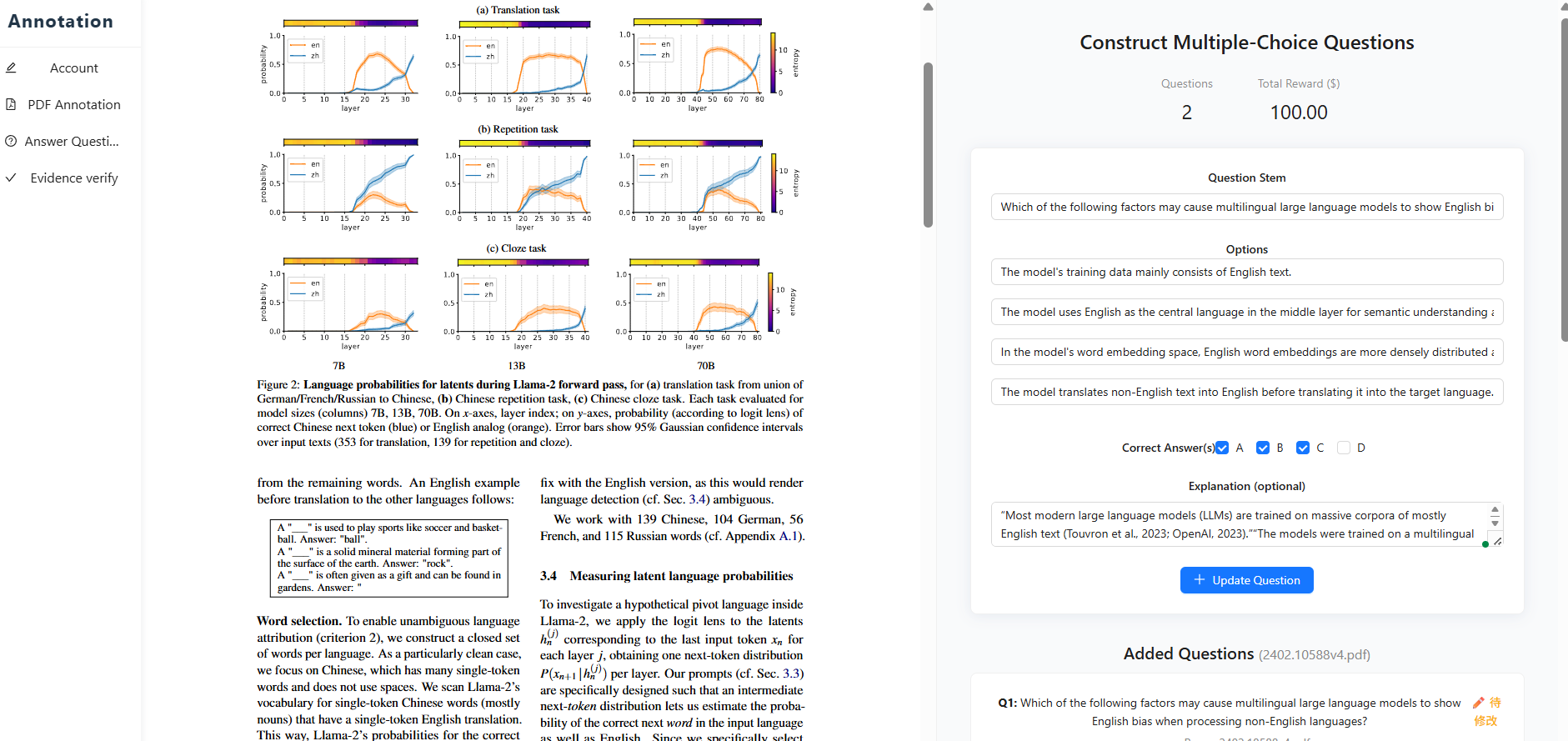}  
  \caption{Screenshot of the QA pair annotation page. Annotators can revise the questions they have created and resubmit them.}
  \label{fig13}
\end{figure*}

\begin{figure*}[htbp]
  \centering
  \includegraphics[width=\textwidth]{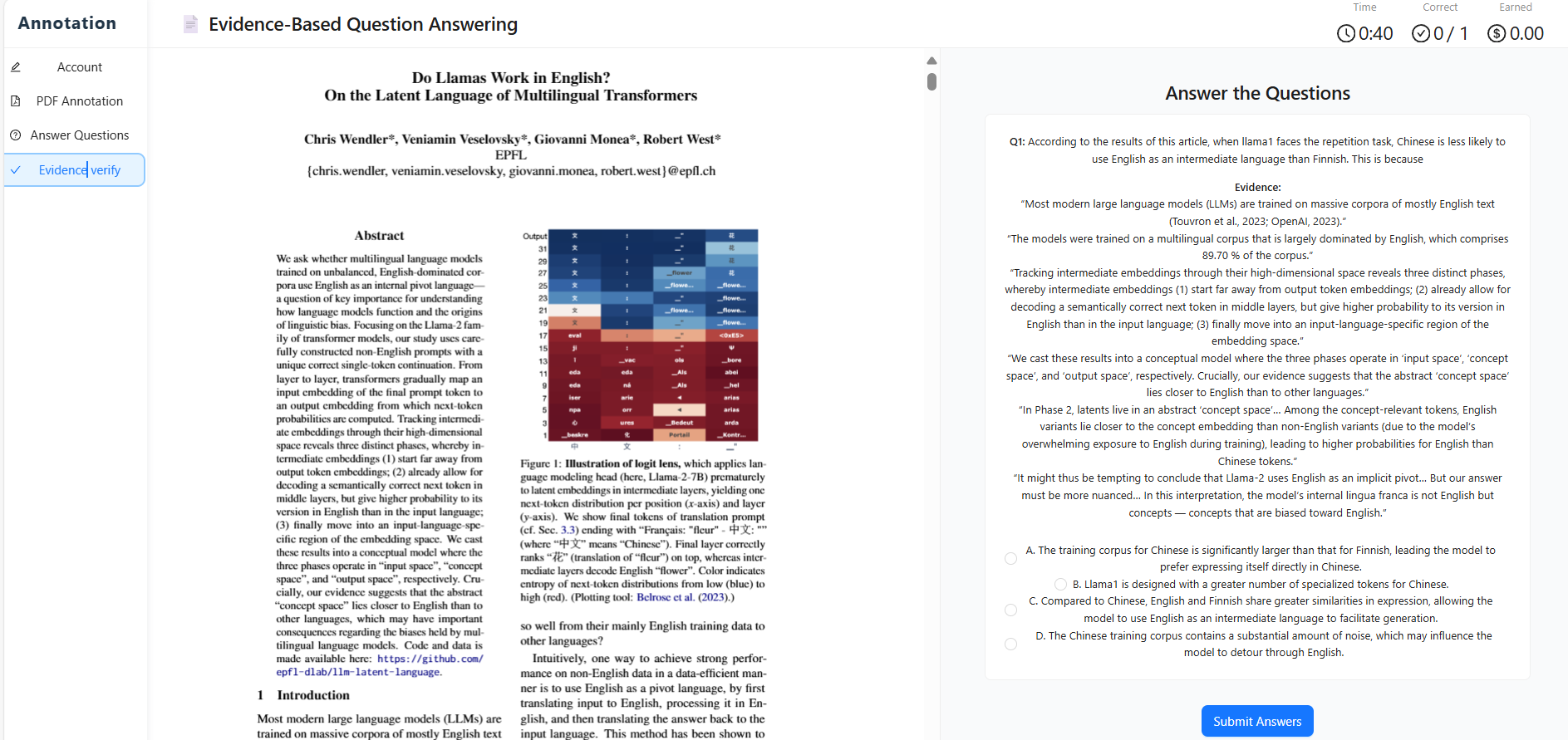}  
  \caption{Screenshot of the evidence verification page. On this page, annotators are provided with a paper, a question, and the corresponding evidence for the question. They are then asked to judge whether the evidence supports the question. Rewards are given based on the correctness of their judgment.}
  \label{fig14}  
\end{figure*}

\begin{figure*}[htbp]
  \centering
  \includegraphics[width=1\textwidth]{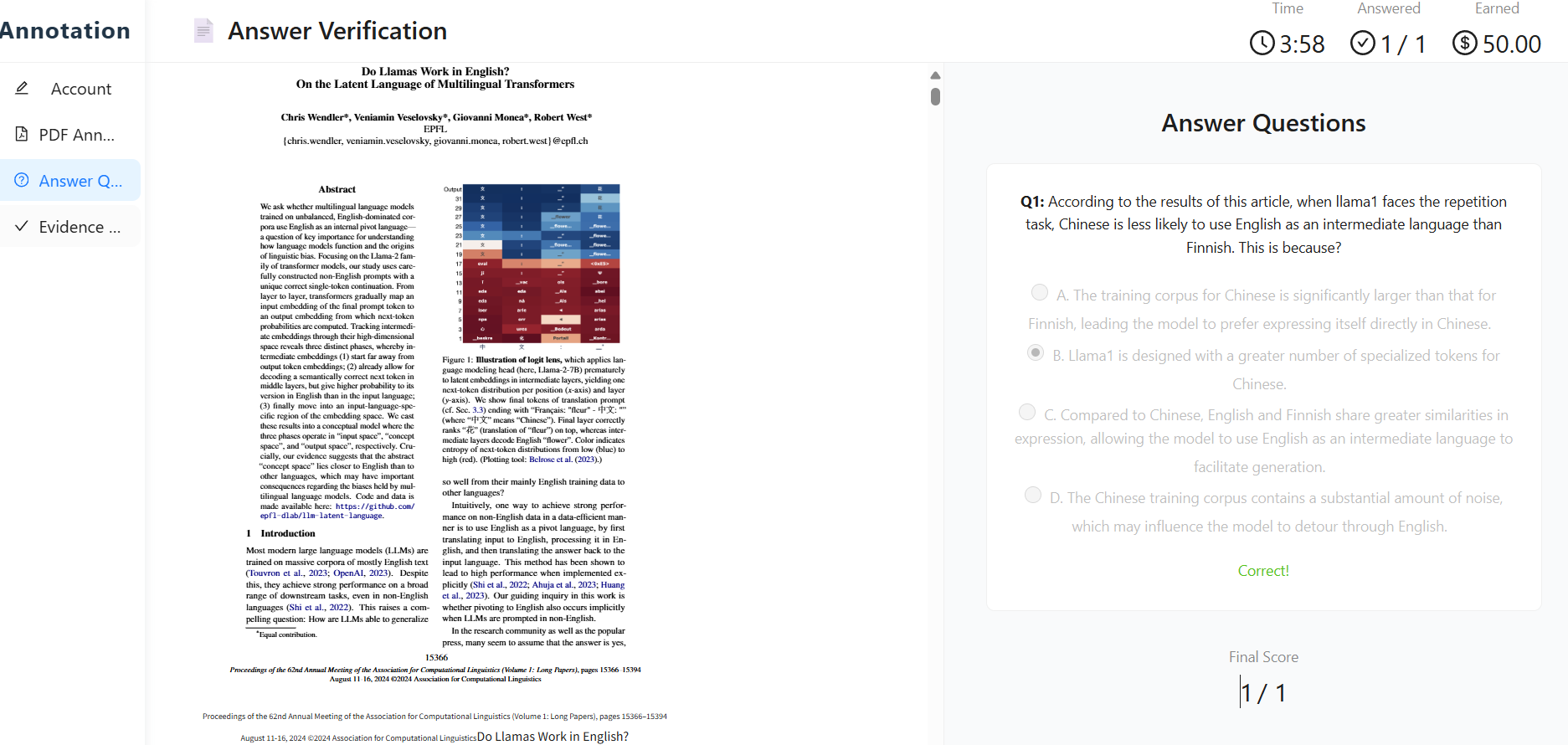}  
  \caption{Screenshot of the answer verification page. \emph{Answer Verifiers} are tasked with answering questions based on the content of the provided paper. During this process, a timer tracks the duration, and rewards are calculated for the \emph{Answer Verifiers} based on both the accuracy of their answers and the response time.}
  \label{fig15}  
\end{figure*}

Our annotation platform is designed to support the construction pipeline for \textsc{ElaipBench}. As shown in Figure~\ref{fig8}, annotators begin by logging into the system, where they can upload papers and track their annotation progress, including the number of QA pairs submitted and the corresponding rewards.
Once a paper is uploaded, \emph{Question Writers} can browse its contents and start creating QA pairs along with the associated supporting evidence on the right side of the interface (Figure~\ref{fig9}). The input interface allows \emph{Question Writers} to write and submit their QA pairs in a structured format (Figure~\ref{fig10}). After submission, they can view their previous submissions and accumulated earnings (Figure~\ref{fig11}).
Submitted QA pairs then undergo an evidence verification stage, where another \emph{Evidence Verifier} is assigned to judge whether the provided evidence adequately supports the question. Annotators can check the verification status of each question (Figure~\ref{fig12}). If the evidence is deemed insufficient or incorrect, the original \emph{Question Writer} is required to revise the QA pair and resubmit it (Figure~\ref{fig13}).
The evidence verification stage (Figure~\ref{fig14}) presents the verifier with the original paper, the question, and the proposed evidence. The verifier must analyze whether the evidence logically supports the question. If the verification is successful, the QA pair proceeds to the subsequent answer verification stage.
In the answer verification stage (Figure~\ref{fig15}), a designated \emph{Answer Verifier} is tasked with answering the question based on the content of the provided paper. A timer runs during the answering process, and both the accuracy and response time are used to calculate the verifier’s reward. The outcome also determines whether the original question writer receives a performance bonus.

\clearpage

\subsection{Benchmark Statistics}
\label{datas}

\begin{figure}[htbp]
  \centering
  \includegraphics[width=\linewidth]{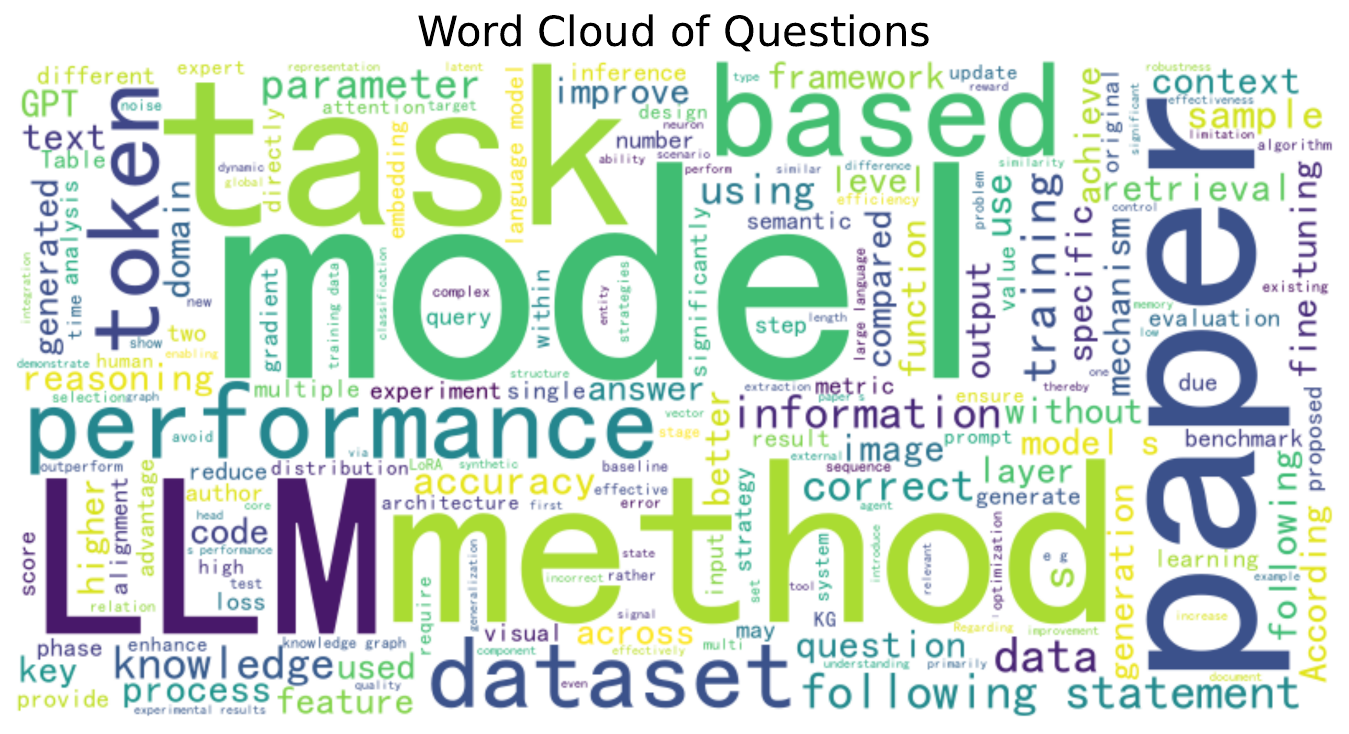}  
  \caption{The word cloud visualization of questions in \textsc{ElaipBench}.}
  \label{fig16}  
\end{figure}

\begin{figure}[htbp]
  \centering
  \includegraphics[width=\linewidth]{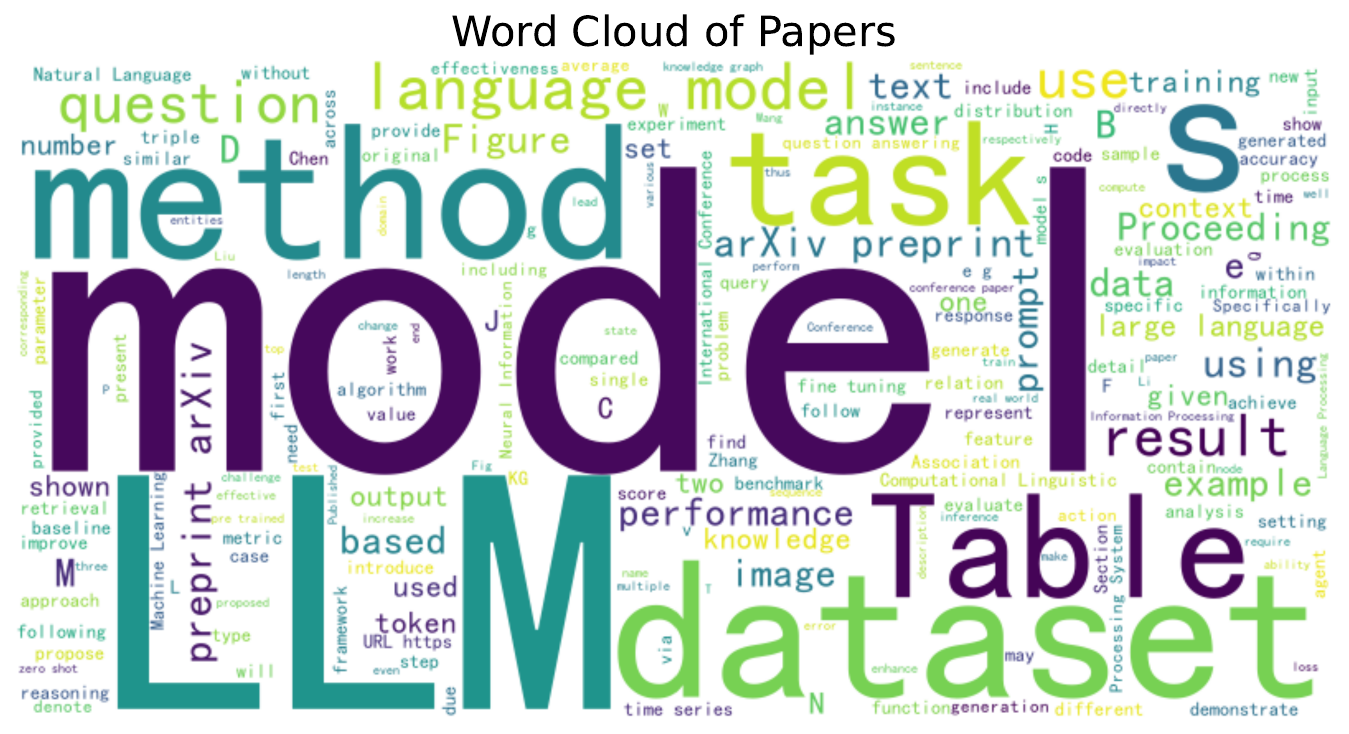}  
  \caption{The word cloud visualization of papers in \textsc{ElaipBench}.}
  \label{fig17}  
\end{figure}

In \textsc{ElaipBench}, the answer options A, B, C, and D account for 26.45\%, 26.10\%, 24.20\%, and 23.25\% of all questions, respectively, indicating a relatively balanced distribution among the four choices.
On average, each article corresponds to 0.64 SA-MCQs and 2.31 MA-MCQs, with 76\% of the papers containing at least one question classified as \textbf{Hard} difficulty. Among all questions, the shortest contains 29 tokens and the longest 492 tokens; across all source papers, the shortest document spans 5,479 tokens, while the longest reaches 31,333 tokens. 
We visualize the word clouds of questions and papers from \textsc{ElaipBench} in Figures~\ref{fig16} and \ref{fig17}. These figures indicate that the benchmark contains a diverse vocabulary for describing academic content.

\begin{table}[htbp]
\centering
\small
\resizebox{\linewidth}{!}{%
\begin{tabular}{lccc}
\toprule
\multirow{2}{*}{\textbf{Discipline}} & \multicolumn{3}{c}{\textbf{Year}} \\ 
\cmidrule{2-4}
 & 2021\textless{} & 2021-2023 & \textgreater{}2023 \\
\midrule
ML & 11 & 9 & 15 \\
CV & 9 & 4 & 11 \\
NLP & 15 & 17 & 46 \\
\bottomrule
\end{tabular}%
}
\caption{Paper statistics.}
\label{tab:papers}
\end{table}

Table~\ref{tab:papers} shows the statistics of the papers. Tables~\ref{tab3} through~\ref{tab8} present representative SA-MCQ and MA-MCQ examples from the three disciplines: ML, CV, and NLP.

\begin{table*}[htbp]
\centering
\begin{tabularx}{\textwidth}{lX}
\toprule
\textbf{Paper Title} & Do Llamas Work in English? On the Latent Language of Multilingual Transformers \\ \midrule
\textbf{Question} & According to the results of this article, when llama1 faces the repetition task, Chinese is less likely to use English as an intermediate language than Finnish. This is because:\newline
A. The training corpus for Chinese is significantly larger than that for Finnish, leading the model to prefer expressing itself directly in Chinese.\newline
B. Llama1 is designed with a greater number of specialized tokens for Chinese.\newline
C. Compared to Chinese, English and Finnish share greater similarities in expression, allowing the model to use English as an intermediate language to facilitate generation.\newline
D. The Chinese training corpus contains a substantial amount of noise, which may influence the model to detour through English. \\ \midrule
\textbf{Answer} & B \\ \bottomrule
\end{tabularx}
\caption{ML Question (SA-MCQ).}
\label{tab3}
\end{table*}

\begin{table*}[htbp]
\centering
\begin{tabularx}{\textwidth}{lX}
\toprule
\textbf{Paper Title} & Machine Unlearning Fails to Remove Data Poisoning Attacks\\ \midrule
\textbf{Question} & Regarding the two main failure reason hypotheses proposed by the authors and how they were validated experimentally, which of the following descriptions is accurate and incisive?\newline
A. Hypothesis 1: Large Displacement refers to the model parameter movement distance required for unloading poisoned samples being significantly greater than that required for unloading an equivalent number of randomly selected clean samples.\newline
B. To validate Hypothesis 2: Orthogonal Subspace, the authors calculated the cosine similarity between the expected offloading direction vector and the direction vector obtained from gradient descent with clean data in a linear regression task and found that this value is close to 1, indicating that the two directions are nearly parallel.\newline
C. When experimentally validating Hypothesis 1, the authors utilized a logistic regression model trained on CIFAR-10 features, as the solution to the convex optimization problem is unique, allowing for clear calculation and comparison of distances between different models.\newline
D. The deeper implication of Hypothesis 2: Orthogonal Subspace is that due to the poisoning effect and normal learning effect being mutually exclusive in the model parameter space, any offloading method that only uses good data to repair the model will ultimately be futile. \\ \midrule
\textbf{Answer} & AC \\ \bottomrule
\end{tabularx}
\caption{ML Question (MA-MCQ).}
\label{tab4}
\end{table*}

\begin{table*}[htbp]
\centering
\begin{tabularx}{\textwidth}{lX}
\toprule
\textbf{Paper Title} & Multimodal ArXiv: A Dataset for Improving Scientific Comprehension of
Large Vision-Language Models \\ \midrule
\textbf{Question} & Which option most accurately summarizes the strategies and findings of the paper regarding dataset construction, model evaluation, and the discussion of current LVLM limitations?\newline
A. The paper finds that although the BLEU-2 score for multi-figure captioning tasks significantly improved for the Qwen-VL-Chat model after fine-tuning on the ArXivCap dataset, GPT-4V demonstrated absolutely leading performance across all vision-to-text tasks, including single-figure, multi-figure, contextualized, and paper title generation, highlighting its powerful zero-shot generalization ability in scientific figure comprehension, far surpassing all open-source models.\newline
B. In the quality control phase of the ArXivCap dataset, to ensure high image quality, the research team not only excluded images with extreme aspect ratios but also filtered out images with the shortest edge shorter than 224 pixels and removed images with pixel numbers exceeding the decompression bombs threshold. These stringent filtering criteria, combined with text cleaning for LaTeX expressions, were key to the dataset's ultimate high quality.\newline
C. The paper conducts a detailed analysis of generated caption error types, where "contextual misinterpretation" is identified as the most prevalent issue for all models, typically stemming from the models' inability to effectively integrate visual information from images with contextual clues from text. "Oversimplification," on the other hand, indicates a tendency for models to generate overly generic descriptions, failing to capture specific details and complex semantics in scientific figures, which reflects the current LVLMs' limitations in deep semantic understanding.\newline
D. All of the above. \\ \midrule
\textbf{Answer} & B \\ \bottomrule
\end{tabularx}
\caption{CV Question (SA-MCQ).}
\label{tab5}
\end{table*}

\begin{table*}[htbp]
\centering
\begin{tabularx}{\textwidth}{lX}
\toprule
\textbf{Paper Title} & SynTab-LLaVA: Enhancing Multimodal Table Understanding with Decoupled Synthesis\\ \midrule
\textbf{Question} & Regarding the construction and characteristics of the SynTab-LLaVA multimodal table understanding synthesis method, which of the following statements are correct?\newline
A. The method decouples the synthesis process into two independent steps: table image rendering and question-answer pair generation, thereby improving efficiency and robustness.\newline
B. Compared with human annotation and traditional MLLM synthesis, the method is more cost-effective, constructing a large-scale dataset with only 200 US dollars.\newline
C. By processing table images generated by Doubao, the LLM can reduce hallucinations and improve the accuracy of Q\&A generation.\newline
D. SynTab-LLaVA adopts a hybrid multi-resolution visual encoder, combining high-resolution and low-resolution image information to capture both local textual content and global structural relationships. \\ \midrule
\textbf{Answer} & AD \\ \bottomrule
\end{tabularx}
\caption{CV Question (MA-MCQ).}
\label{tab6}
\end{table*}

\begin{table*}[htbp]
\centering
\begin{tabularx}{\textwidth}{lX}
\toprule
\textbf{Paper Title} & Exploring the Impact of Table-to-Text Methods on Augmenting LLM-based
Question Answering with Domain Hybrid Data \\ \midrule
\textbf{Question} & Under the Domain-Specific Fine-Tuning framework, according to GPT-4's evaluation criteria, which model in the OPT series is the most sensitive to different Table-to-Text methods?\newline
A. OPT-1.3B\newline
B. OPT-2.7B\newline
C. OPT-6.7B\newline
D. OPT-13B \\ \midrule
\textbf{Answer} & A \\ \bottomrule
\end{tabularx}
\caption{NLP Question (SA-MCQ).}
\label{tab7}
\end{table*}

\begin{table*}[htbp]
\centering
\begin{tabularx}{\textwidth}{lX}
\toprule
\textbf{Paper Title} & Astute RAG: Overcoming Imperfect Retrieval Augmentation and Knowledge Conflicts for Large Language Models\\ \midrule
\textbf{Question} & Which approach most effectively mitigates the negative consequences of flawed information retrieval in RAG systems, as suggested by recent research?\newline
A. A primary approach involves exclusively refining the initial retrieval mechanism through techniques like advanced query expansion and document re-ranking to guarantee the relevance of the provided external knowledge.\newline
B. One compelling approach is the sophisticated integration of the language model's pre-existing internal knowledge with the retrieved external information, employing strategies to identify and resolve potential conflicts in a source-aware manner.\newline
C. A straightforward approach to enhance RAG system resilience focuses on increasing the sheer volume of retrieved documents, assuming that the probability of encountering correct information will proportionally rise with a larger context window.\newline
D. An alternative approach leverages a multi-stage process where the language model first generates potential answers based solely on its internal knowledge, and subsequently uses retrieved documents merely as a validation layer without actively integrating conflicting information. \\ \midrule
\textbf{Answer} & BD \\ \bottomrule
\end{tabularx}
\caption{NLP Question (MA-MCQ).}
\label{tab8}
\end{table*}

\subsection{Evaluation Metric and LLM Settings}
\label{models}

\paragraph{Evaluation Metric.} Our evaluation employs strict accuracy metrics: for MA-MCQ, responses must identify all correct options without errors; SA-MCQ requires exact match.

\paragraph{Prompts.} 

For the base models:

\begin{tcolorbox}[colback=gray!10, colframe=darkgray, title=SA-MCQ Prompt, width=\linewidth]
Please select the correct answer based on the paper content. 
Each question has only 1 correct option. 
Format your response as follows:

The correct answer is boxed \{insert answer here\}.

Examples:

The correct answer is boxed \{A\}

Paper: \{paper\}

Question: \{question\}

\end{tcolorbox}

\begin{tcolorbox}[colback=gray!10, colframe=darkgray, title=MA-MCQ Prompt, width=\linewidth]
Please select all correct answers based on the paper content. 
Each question has 2-3 correct options.
Format your response as follows:

The correct answer is boxed \{insert answer here\}.

Examples:

The correct answer is boxed \{ACD\}

Paper: \{paper\}

Question: \{question\}

\end{tcolorbox}

For the CoT paradigm, we add \textit{``Let's think step by step.''} to the prompt. For the LRMs, the prompts are the same as the base models.

\paragraph{Parameter Settings.} All evaluated LLMs feature context windows of at least 32k tokens, enabling comprehensive coverage of full papers and questions. In the evaluation setup for base models, we configure the generation sampling parameters with temperature=0.1 and max\_new\_tokens=32. For the base models with CoT setting, we employ temperature=0.1 and max\_new\_tokens=4096. The same parameter configuration is adopted for LRMs to accommodate their extended reasoning capabilities. This standardized parameter setup ensures fair comparison across different model configurations while allowing sufficient generation length for complex reasoning tasks.

\begin{figure}[htbp]
  \centering
  \includegraphics[width=\linewidth]{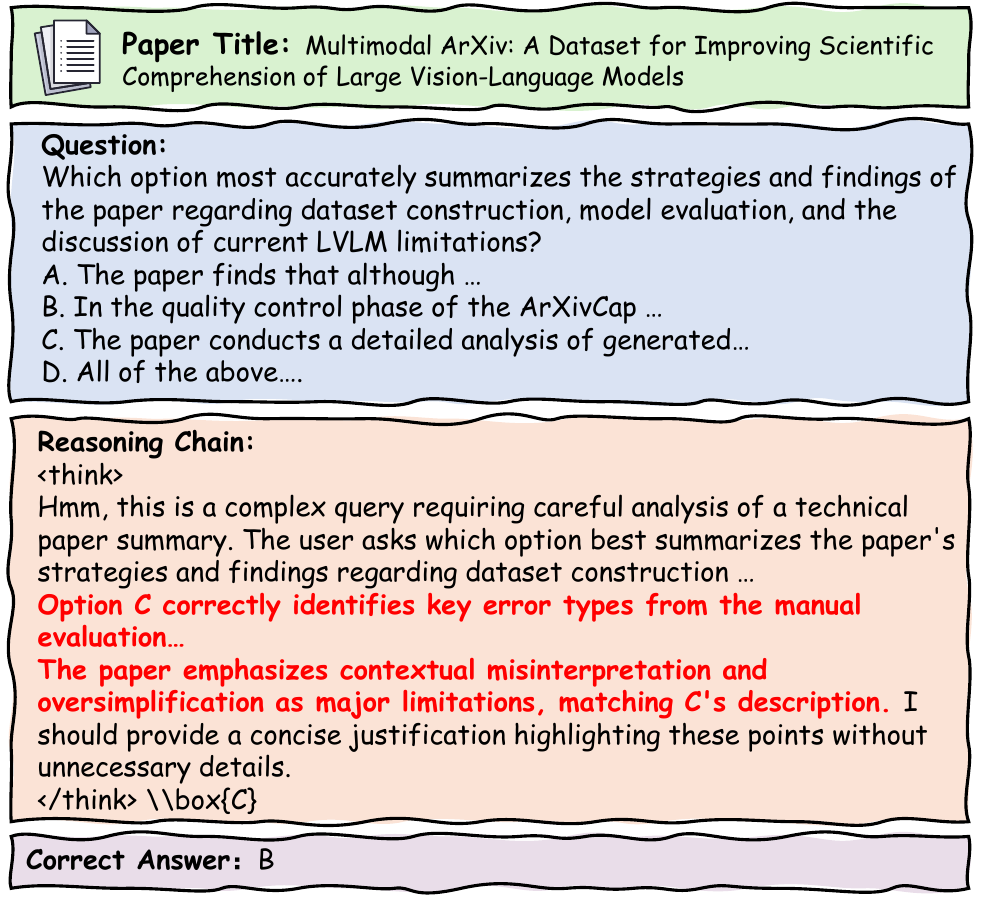}
  \caption{An example of \textbf{Analytical Error} from DeepSeek-R1, where the LLM commits to an incorrect answer from the outset.}
  \label{fig18}
\end{figure}

\begin{figure}[htbp]
  \centering
  \includegraphics[width=\linewidth]{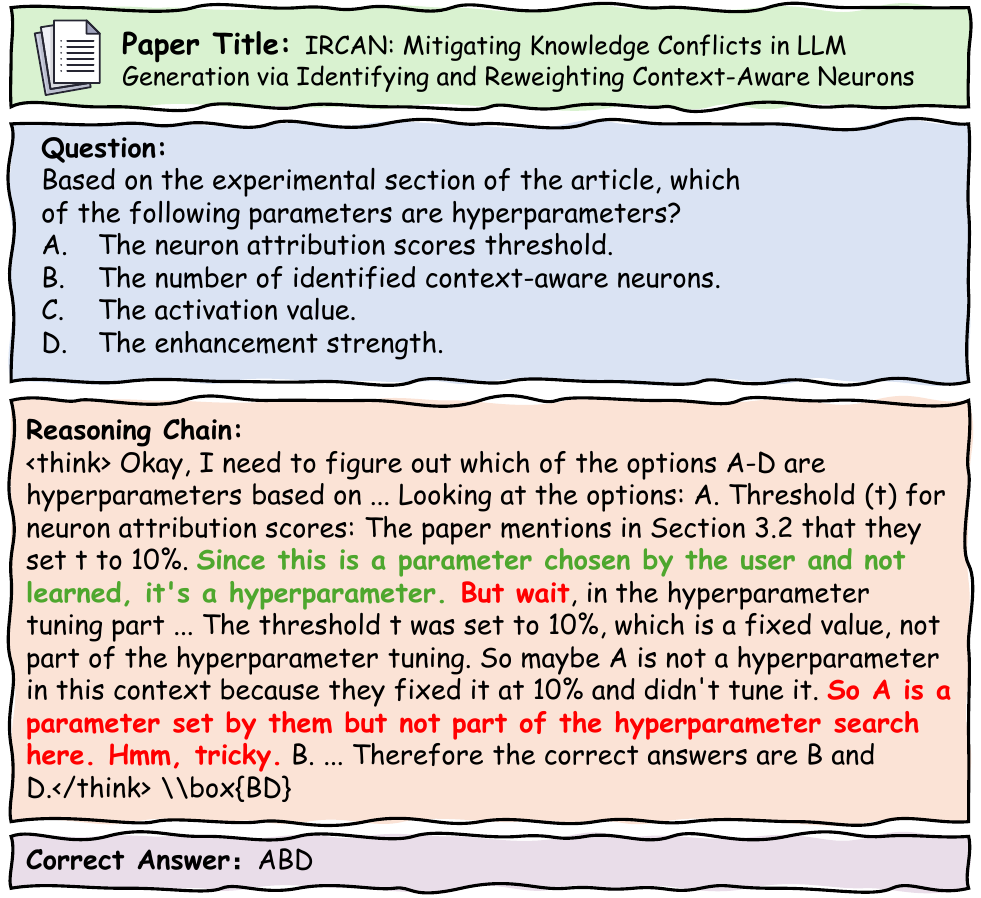}
  \caption{An example of \textbf{Harmful Verification} from QwQ-32B, where the LLM first identifies the correct answer but then incorrectly rejects it during reasoning, ultimately selecting a wrong alternative.}
  \label{fig19}
\end{figure}

\begin{figure}[htbp]
  \centering
  \includegraphics[width=\linewidth]{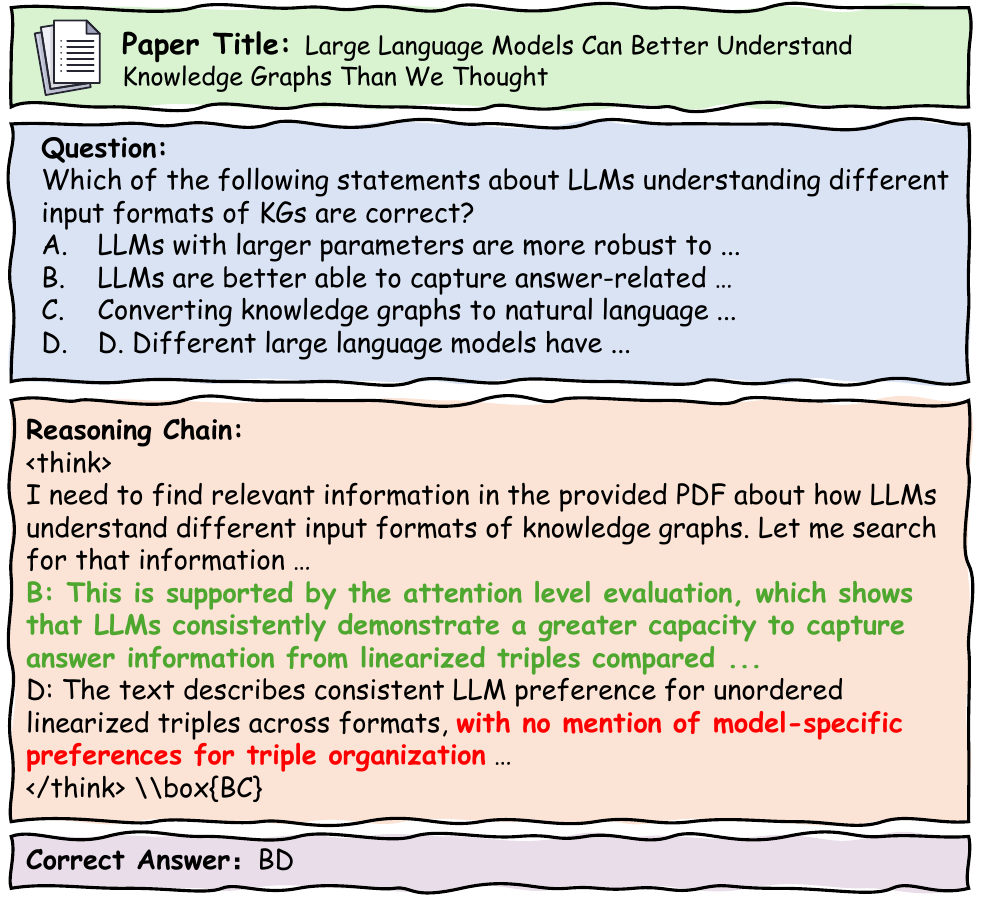}
  \caption{An example of \textbf{Overconfidence} from Claude-3.7-Sonnet-thinking, where the LLM fails to carefully review the paper, assumes that no content relevant to the options is present, and ultimately selects an incorrect choice.}
  \label{fig20}
\end{figure}

\subsection{Error and Efficiency Analysis}
\label{error_analysis}

Figure~\ref{fig18} presents a case of \textbf{Analytical Error}, in which the model commits an incorrect analysis at the outset and prematurely settles on a flawed conclusion. Figure~\ref{fig19} illustrates an instance of \textbf{Harmful Verification}, where the model expends a substantial number of tokens during the verification phase yet ultimately revises a correct answer into an incorrect one. Figure~\ref{fig20} demonstrates \textbf{Overconfidence}, in which the model disregards the specific context or arguments presented in the source paper and instead relies on its internal priors to directly select an option, bypassing contextual reasoning.

\begin{figure}[htbp]
  \centering
  \includegraphics[width=\linewidth]{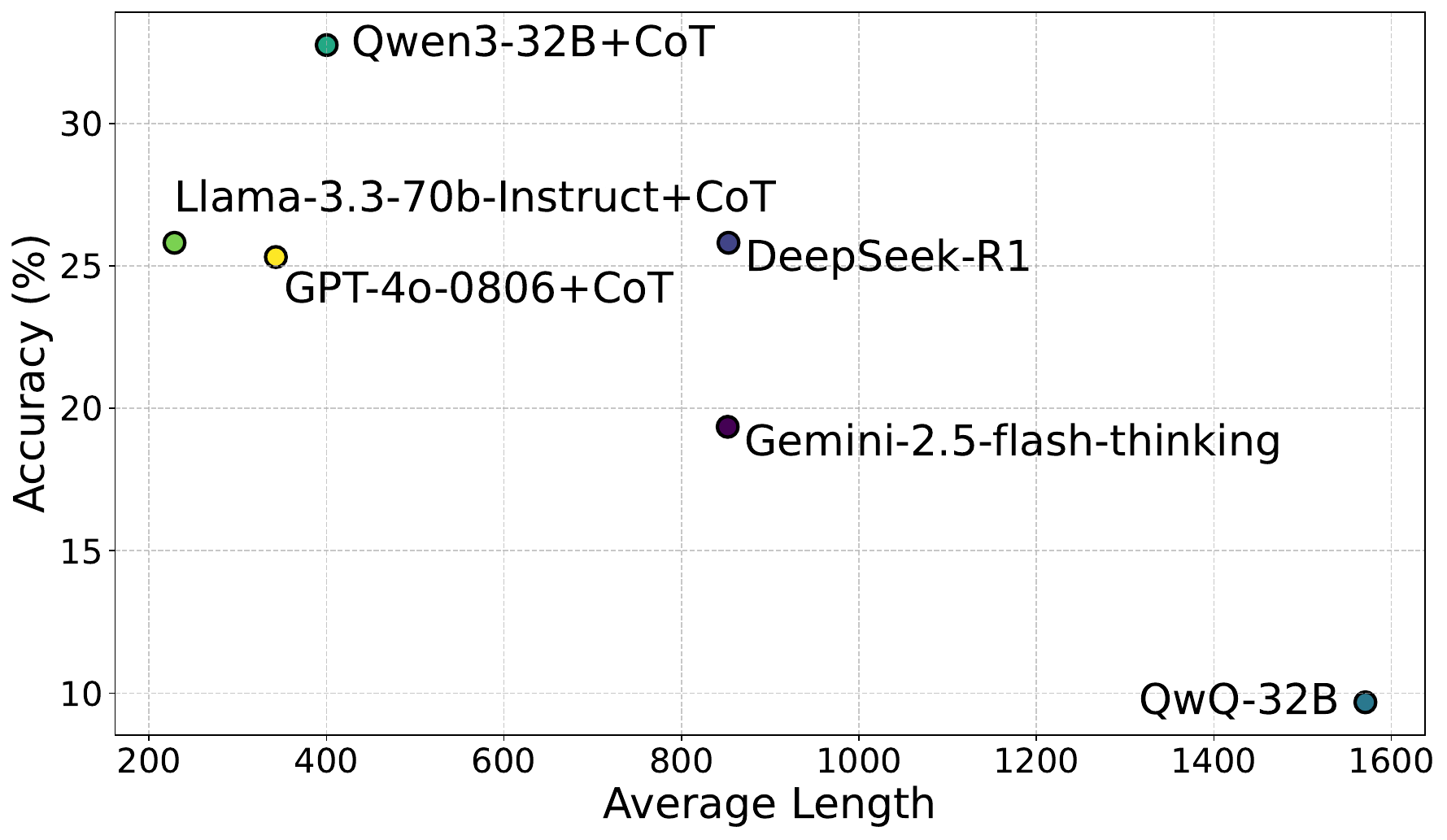}
  \caption{Completion Tokens vs.Performance.}
  \label{fig21}
\end{figure}

Figure~\ref{fig21} illustrates the relationship between average reasoning length and accuracy across different LLMs. Although QwQ-32B generates the longest reasoning chains, its accuracy does not improve with increased length—in fact, it achieves the lowest performance among the models. DeepSeek-R1 attains the highest accuracy among the three LRMs while maintaining a relatively short average reasoning chain, demonstrating superior reasoning efficiency. Nevertheless, its performance remains significantly lower than that of Qwen3-32B+CoT, suggesting that, in many models, longer reasoning chains do not necessarily lead to better performance and may even be detrimental.

\subsection{RAG Settings}
\label{rag_settings}

\subsubsection{Hyper-parameter Setting}

For the dense passage retrieval in the intra-paper setting, we employ the {\tt BGE-m3}~\cite{chen2024m3} encoder with its official checkpoint.\footnote{www.huggingface.co/BAAI/bge-m3}
Each segmented passage is embedded into a 1024-dimensional vector space using the encoder's default configuration. During retrieval, we adopt a cosine similarity metric to measure relevance between a question embedding and candidate passage embeddings. We set the {\tt top\_k} parameter to $5$, returning the five highest-scoring passages per query.

For the BM25-based retrieval, we utilize the {\tt rank\_bm25} implementation with the default hyper-parameters $k_1=1.5$ and $b=0.75$. Queries are tokenized using the same sentence boundary segmentation as in the DPR setup, ensuring consistency.

In the web-based retrieval setting, we interact with the Google Custom Search API, forming queries by concatenating the question and paper title. We restrict the API to return at most $10$ results, from which we select the top-$5$ after relevance filtering. All retrieved HTML documents are converted to plain text using the {\tt trafilatura} library and truncated to the first $4096$ tokens for efficiency.

\subsubsection{RAG Configuration}

In both retrieval paradigms, retrieved passages or web documents are treated as external knowledge sources. Before being fed into the LLM, these snippets undergo the following preprocessing steps:

\begin{itemize}
\item \textbf{Text Normalization}: Remove excessive whitespace, HTML tags, and non-textual artifacts.
\item \textbf{Sentence Trimming}: Ensure that each snippet begins and ends at complete sentence boundaries to preserve semantic integrity.
\item \textbf{Token Budgeting}: Maintain a combined context length (paper + knowledge snippets) within 32,768 tokens to fit the LLM's input limits; in cases exceeding this limit, we perform a round-robin truncation over retrieved items.
\item \textbf{Ordering}: Within each query, snippets are ordered in descending retrieval score before concatenation into the prompt.
\end{itemize}

\subsubsection{RAG Prompts}

In prompts, we explicitly instruct the LLM to treat this content as relevant context and to consult it when formulating its response. The prompt is structured as follows:

\begin{tcolorbox}[colback=gray!10, colframe=darkgray, title=SA-MCQ Prompt, width=\linewidth, breakable]
Please select the correct answer based on the content of the paper and the retrieved knowledge snippets provided below.  
Each question has only 1 correct option. 
Format your response as follows:

The correct answer is boxed \{insert answer here\}.

Examples:

The correct answer is boxed \{A\}

Paper: \{paper\}

Knowledge snippets: \{retrieved knowledge\}  

Question: \{question\}

\end{tcolorbox}

\begin{tcolorbox}[colback=gray!10, colframe=darkgray, title=MA-MCQ Prompt, width=\linewidth, breakable]
Please select all correct answers based on the content of the paper and the retrieved knowledge snippets provided below. 
Each question has 2-3 correct options.
Format your response as follows:

The correct answer is boxed \{insert answer here\}.

Examples:

The correct answer is boxed \{ACD\}

Paper: \{paper\}

Knowledge snippets: \{retrieved knowledge\}  

Question: \{question\}

\end{tcolorbox}

\end{document}